\documentclass[runningheads]{llncs}

\usepackage{eccv}

\usepackage{eccvabbrv}

\usepackage{graphicx}
\usepackage{booktabs}

\usepackage[accsupp]{axessibility}  

\usepackage{hyperref}

\usepackage{orcidlink}
\usepackage{multirow}
\usepackage{siunitx}

\begin{document}

\title{A Benchmark for Heterogeneous Stereo Deblurring with Physically- and Epipolar-constrained Cross Attention}

\titlerunning{A Benchmark for Heterogeneous Stereo
Deblurring with PECA}

\makeatletter
\renewcommand{\@fnsymbol}[1]{\ensuremath{\ifcase#1\or *\or \dagger\or
  \ddagger\or \mathsection\or \mathparagraph\or \|\else\@ctrerr\fi}}
\makeatother
\newcommand{\samethanks}[1][\value{footnote}]{\footnotemark[#1]}

\author{Hoju Shin\inst{1}\thanks{Equal contribution.}\orcidlink{0009-0003-0680-523X}
\and Jiah Kim\inst{1}\samethanks\orcidlink{0009-0006-0387-4098}
\and Seung-Wook Kim\inst{1}\thanks{Corresponding author.}\orcidlink{0000-0002-6004-4086}
\and Seowon Ji\inst{2}\samethanks\orcidlink{0000-0001-8700-8440}}

\authorrunning{H.~Shin et al.}

\institute{Department of Intelligent Robot Engineering, Pukyong National University, Busan, Republic of Korea\\ 
\and Department of Computer Science and Engineering, Konkuk University, Seoul, Republic of Korea\\
\email{\{hjshin, whatime08\}@pukyong.ac.kr, swkim@pknu.ac.kr, seowonji@konkuk.ac.kr} }

\maketitle

\begin{abstract}
 Modern stereo-capable smartphones enable immersive XR content capture. However, hardware heterogeneity across camera modules often causes severe asymmetric blur artifacts.
 Existing methods and benchmarks largely assume homogeneous stereo setups and therefore do not explicitly address such asymmetric degradation. 
 To bridge this gap, we present a dedicated framework for heterogeneous stereo deblurring. 
 First, we introduce the heterogeneous stereo deblurring (HSD) dataset, constructed from real smartphone stereo captures via multi-frame integration. 
 Second, we propose physically- and epipolar-constrained cross attention (PECA), a lightweight module that restricts cross-view matching to an epipolar search window bounded by a optics-derived disparity upper bound. 
 By enforcing physically valid disparity constraints, PECA enables efficient and reliable cross-view feature fusion. 
 Moreover, our confidence-weighted attention with residual fusion emphasizes cross-guided deblurring when correspondences are reliable, while naturally falling back to self-deblurring in occluded or unreliable regions. 
 PECA is architecture-agnostic and consistently improves CNN-, Transformer-, and NAFNet-based baselines. 
 Extensive experiments on HSD show that PECA-enhanced models achieve improved restoration performance with favorable efficiency. The dataset and source code are publicly released at \url{https://github.com/shinhoju/PECA}.

\end{abstract}

\section{Introduction}
\label{sec:intro}

Modern smartphones have evolved into sophisticated imaging devices by adopting heterogeneous multi-camera systems as a standard configuration.
These systems generally combine a primary wide camera with auxiliary modules, such as ultra-wide or telephoto lenses~\cite{delbracio2021mobile}.
Beyond enhancing single-view photography, this configuration enables stereoscopic capture and other emerging 3D applications that require synchronized stereo observations to reconstruct content for binocular displays and head-mounted devices~\cite{kim2021generation}.

Although this multi-camera configuration supports stereoscopic capture, modern smartphones are not built as calibrated, homogeneous stereo rigs. 
Instead, they integrate heterogeneous camera modules, resulting in systematic asymmetry in image quality across views.
In common devices, the primary wide camera typically benefits from a bright aperture and optical image stabilization~(OIS), whereas the ultra-wide camera is often constrained by a narrower aperture and lacks dedicated stabilization.
These camera properties make the ultra-wide stream more susceptible to motion.
Consequently, stereoscopic captures frequently exhibit a sharp wide view alongside a blur-degraded ultra-wide view, reflecting hardware-induced blur asymmetry.

\begin{figure}[tb]
  \centering
  \includegraphics[width=0.9\linewidth]{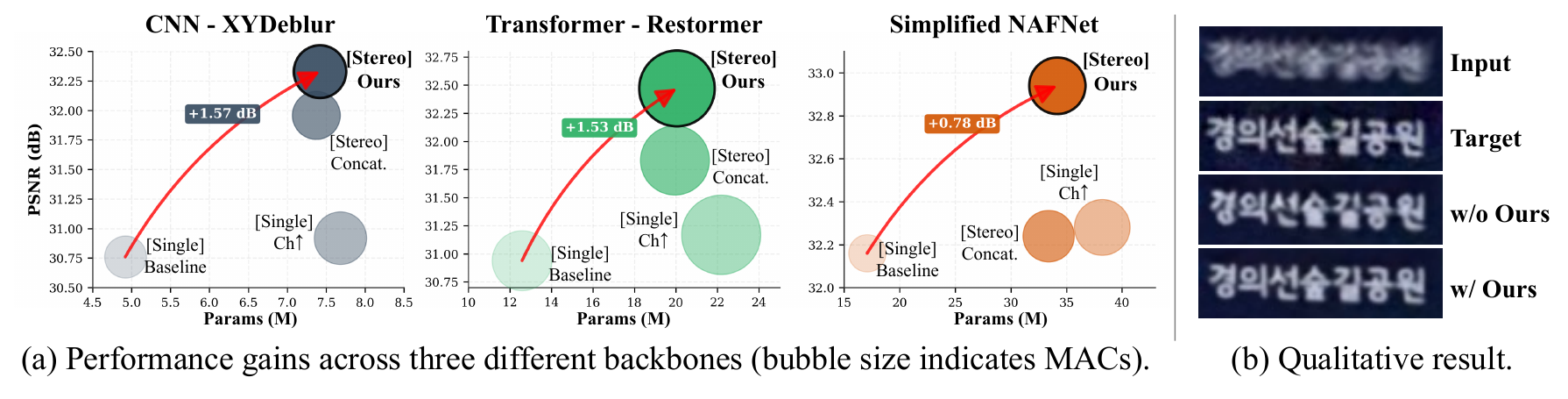}
  \vspace{-3mm}
  \caption{Overview of PECA performance and visual quality on the HSD dataset. (a) Accuracy-efficiency trade-off across three representative backbones (XYDeblur, Restormer, and NAFNet). (b) Qualitative restoration results on a challenging scene.}
  \label{fig:teaser}
  \vspace{-3mm}
\end{figure}

The use of synchronized stereo cameras naturally suggests using a sharper view to restore a degraded one. 
However, most existing restoration approaches are not designed with asymmetric hardware degradation as the primary objective.
Many stereo restoration methods are developed under homogeneous camera assumptions~\cite{wang2019learning, wang2021symmetric, zou2023cross, sellent2016stereo, pan2017simultaneous, pan2019joint, zhou2019davanet}, where both views share comparable imaging characteristics and restoration is formulated symmetrically. 
In such settings, cross-view correspondence is mainly exploited to resolve geometric misalignment or transfer details between similarly degraded views. 
This symmetric formulation does not reflect heterogeneous stereo-camera systems, where one view often serves as a high-quality reference while the other suffers from systematic blur.

Recent works have begun exploring heterogeneous multi-camera systems~\cite{wang2021dual, yue2024kedusr, kang2025dual, lee2022reference, kim2023efficient, zou2025refvsr++, xiaoasymmetric, rim2024deep, shekarforoush2023dual, lin2025learning}. 
Although these approaches exploit cross-camera complementarity, many rely on unrectified stereo data and global dense correspondence estimation or iterative alignment, which may incur significant computational overhead and can be sensitive to sensor heterogeneity. 
Moreover, asymmetric blur is rarely formulated as the primary restoration target under device-level constraints.
Consequently, there remains a lack of a dedicated framework and benchmark that directly deal with hardware-induced blur asymmetry under rectified heterogeneous stereo settings.

In this work, we study heterogeneous stereo deblurring in practical smartphone settings, where hardware-induced blur asymmetry across views is the primary restoration challenge.
We introduce a real-world benchmark, the heterogeneous stereo deblurring (HSD) dataset, constructed from synchronized wide and ultra-wide video sequences with device-provided geometric rectification to reflect real mobile pipelines.
In HSD, the wide view serves as a sharp reference, while the ultra-wide blur is approximated by temporal integration to emulate exposure-induced degradation. 
This design decouples geometric alignment from blur restoration, enabling controlled evaluation under practical device environments.
To establish a physically grounded and efficient reference model for HSD, we propose physically- and epipolar-constrained cross attention (PECA), a lightweight cross-view fusion module.
Leveraging rectified geometry and an optics-based disparity upper bound derived from the minimum focus distance of the stereo pair, PECA restricts cross-view attention to a localized 1D window along the epipolar line. 
The disparity bound determines the maximum physically feasible correspondence, while rectified stereo geometry further imposes a directional constraint on the search. 
Together, these constraints confine attention to a physically valid and computationally efficient region.
The retrieved cross-view features are then integrated into the target stream via residual fusion, which preserves view-specific characteristics. 
As illustrated in Fig.~\ref{fig:teaser}, PECA consistently improves performance on the HSD dataset across diverse backbone architectures. 
Our contributions are summarized as follows:
\begin{enumerate}
    \item We introduce \textbf{HSD}, a real-world benchmark of synchronized wide/ultra-wide smartphone stereo captures with device-provided rectification, designed to study hardware-induced asymmetric blur under practical mobile camera pipelines.
    \item We propose \textbf{PECA}, a lightweight cross-view fusion module that constraints attention to a directional 1D epipolar window whose extent is bounded based on optics, enabling efficient and robust cross-view feature retrieval without global dense matching.
    \item We demonstrate the \textbf{backbone-agnostic effectiveness} of PECA across CNN-, Transformer-, and modern simplified NAFNet-based models, consistently improving restoration quality with favorable efficiency on HSD.
\end{enumerate}

\section{Related Work}
\subsection{Single-Image Deblurring}
Single-image deblurring has been widely studied, evolving from kernel-estimation approaches~\cite{zhang2025survey} to modern learning-based models such as CNNs, Transformers, and activation-free networks~\cite{cho2021rethinking, ji2022xydeblur, chen2022simple, kong2023efficient, liu2023real, liu2025xyscannet}. 
Representative datasets such as GoPro~\cite{nah2017deep}, REDS~\cite{nah2019ntire}, RealBlur~\cite{rim2020real}, and HIDE~\cite{shen2019human} have played a crucial role in advancing single-view deblurring by providing paired blurry-sharp data under diverse motion patterns and camera settings.

However, these methods operate on a single observation and cannot exploit cross-view redundancy available in multi-camera systems. Consequently, they are not designed to address asymmetric degradation across synchronized views, nor are existing single-image benchmarks suitable for evaluating heterogeneous stereo deblurring.

\begin{figure}[tb]
  \centering
  \includegraphics[width=\linewidth]{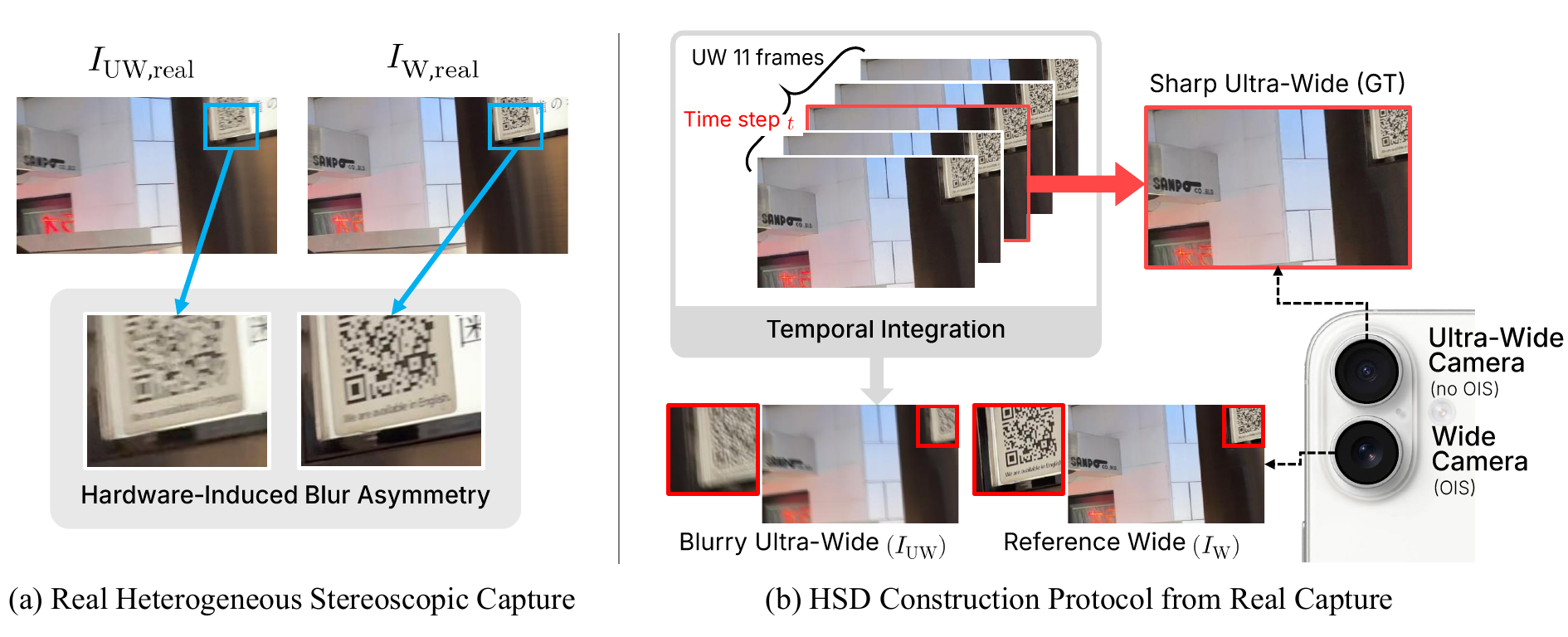}
  \vspace{-5mm}
  \caption{Overview of the HSD benchmark construction from real heterogeneous stereo capture.
  (a) Real smartphone stereoscopic capture, where hardware-induced blur asymmetry naturally arises between wide and ultra-wide modules.
  (b) Construction protocol applied to real synchronized stereo sequences. 
  }
  \vspace{-3mm}
  \label{fig:DB}
\end{figure}

\vspace{-2mm}
\subsection{Stereo Image Restoration}

\subsubsection{Homogeneous stereo restoration}
Stereo restoration methods have explored cross-view correspondence to improve reconstruction quality~\cite{wang2019learning, wang2021symmetric, zou2023cross}. 
In stereo deblurring, prior works leverage scene flow estimation, joint optimization, or disparity-aware feature aggregation to restore latent sharp images~\cite{sellent2016stereo, pan2017simultaneous, pan2019joint, zhou2019davanet}. 
Related approaches in stereo super-resolution (SR) similarly exploit cross-view correspondences to transfer fine-grained details.

These methods are typically developed under homogeneous camera assumptions, where both views share similar sensor characteristics and image quality. Restoration is therefore formulated symmetrically, treating each view as both source and target. While effective in controlled homogeneous stereo setups, this assumption does not hold in heterogeneous smartphone systems, where auxiliary cameras often exhibit systematic quality degradation.

\vspace{-3mm}
\subsubsection{Heterogeneous stereo restoration}
Recent works have explored heterogeneous multi-camera systems, leveraging cross-camera complementarity for SR and deblurring~\cite{wang2021dual, yue2024kedusr, kang2025dual, lee2022reference, kim2023efficient, zou2025refvsr++, xiaoasymmetric, shekarforoush2023dual, lin2025learning}.
In mobile imaging, hybrid-camera deblurring has also been studied by combining complementary captures across smartphone modules and introducing datasets for evaluation~\cite{rim2024deep}.
These lines of work highlight the potential of cross-camera information in asymmetric settings.

However, many approaches operate on unrectified stereo data and rely on global dense correspondence estimation or iterative alignment, incurring significant computational overhead and sensitivity to sensor heterogeneity. Although device-provided rectification reduces correspondence search to a 1D epipolar space, this structural prior remains under-exploited for efficient asymmetric restoration. Furthermore, the lack of a dedicated benchmark tailored to rectified heterogeneous stereo with controlled blur asymmetry has limited systematic evaluation under realistic device pipelines.

\vspace{-3mm}
\subsubsection{Epipolar-aware attention}
In stereo SR, attention-based stereo fusion has been adopted to enhance cross-view feature interaction. Some methods such as PASSRnet~\cite{wang2019learning}, iPASSR~\cite{wang2021symmetric}, and NAFSSR~\cite{chu2022nafssr} restrict correspondence search to horizontal epipolar lines, reducing complexity compared to global 2D attention.

Nevertheless, these approaches typically search across the full image width without explicit physical constraints. Without camera-optics-derived disparity bounds, attention may aggregate features from geometrically implausible regions along the epipolar line. Moreover, correspondence search is often performed symmetrically along the scanline, ignoring the physically valid disparity direction imposed by rectified stereo geometry. 
In contrast, our work incorporates an optics-based disparity upper bound and a directional epipolar constraint to confine cross-view matching to a physically plausible and computationally efficient search region.

\vspace{-2mm}
\section{HSD Benchmark Construction}
\label{hsd}
\vspace{-1mm}

As shown in Fig.~\ref{fig:DB}(a), heterogeneous smartphone cameras often exhibit systematic blur asymmetry between wide and ultra-wide modules.
To study this phenomenon, we construct a real-world benchmark for heterogeneous stereo deblurring from synchronized stereo captures.
The benchmark is built from real heterogeneous stereo capture, preserving device-level characteristics while introducing controlled exposure simulation.
It is designed to (i) reflect device-level camera processing used in commodity smartphones, (ii) approximate physically plausible motion blur, and (iii) avoid trivial or near-static cases that would otherwise obscure the asymmetric deblurring problem.

\vspace{-2mm}
\subsection{Data Acquisition and Camera Configuration}
\vspace{-1mm}
All data are captured using the stereoscopic capture mode of commercial smartphones (e.g., iPhone 15 Pro Max, iPhone 16 series, and iPhone 17) equipped with heterogeneous stereo systems.
Each recording provides synchronized wide and ultra-wide video streams observing the same scene.
The captured videos are rectified through the native imaging pipeline, which dynamically compensates for variations in focal length and principal points induced by the auto focus~(AF) and OIS, producing epipolar-aligned stereo frames. 
We directly adopt the device-provided rectified outputs without additional calibration to reflect practical deployment scenarios.
The final collection contains diverse indoor and outdoor environments with varying lighting and depth distributions.

\vspace{-2mm}
\subsection{Blur Generation via Temporal Integration}
\label{main:db}
\vspace{-1mm}
A key challenge in stereo deblurring benchmarks is obtaining blur–sharp pairs that reflect real-world image formation. 
In heterogeneous smartphone systems, motion blur in the ultra-wide stream primarily arises from longer effective exposure times and weaker stabilization relative to the primary wide camera.
To model this exposure-induced degradation, we adopt the temporal integration protocol established by Nah \textit{et al.}~\cite{nah2017deep} while adapting it to the constraints of a synchronized stereo capture system.

Unlike conventional benchmarks that rely on high-speed cameras (e.g., 240 fps), such synchronized capture is impractical for commercial smartphones.
Instead, we adopt a scene-diversity-centric acquisition strategy.
We collect a large set of 384 scenes with controlled and stable camera motion, ensuring moderate inter-frame displacement and avoiding unnatural artifacts during temporal integration.

As shown in Fig.~\ref{fig:DB}(b), for each sequence, we generate blurred ultra-wide frames by averaging consecutive frames centered at a target time step $t$. This accumulation mimics a longer exposure interval in which a sensor integrates light over time. 
Since our raw sequences are captured with stable motion, this integration results in a smooth, continuous blur that faithfully approximates real-world degradation without the discrete step artifacts typical of low-FPS averaging with fast motion. 
Because the integration follows real handheld camera trajectories, the resulting blur is depth-dependent and spatially varying rather than uniform.
The central frame is treated as the ground truth, while the synchronized wide-view frame at the same timestamp serves as the sharp cross-view reference.
We do not explicitly model the rolling shutter effect, as this would require assumptions about device-specific readout patterns and ISP patterns. Consequently, the HSD dataset does not fully account for rolling-shutter distortions.
To mitigate noise artifacts while maintaining tractable computational complexity, we downscale the synthesized ultra-wide blurry frames as well as the recorded sharp ultra-wide and wide frames from $1080\times1920$ to $720\times1280$ using bicubic interpolation. 

\vspace{-2mm}
\subsection{Dataset Curation and Partitioning}
\vspace{-1mm}
To ensure that the benchmark presents a non-trivial challenge and maintains evaluation validity, we curate the dataset before final partitioning.
Since temporal integration does not always guarantee sufficiently severe degradation, we filter out near-static samples where integration yields negligible blur.
Specifically, we compute the peak signal-to-noise ratio (PSNR) between each integrated blurry image and its corresponding sharp ultra-wide frame, and discard pairs whose PSNR exceeds 33dB.
This filtering step ensures that the benchmark emphasizes scenarios where asymmetric deblurring provides meaningful gains.

After curation, we split the 384 sequences at the sequence level to prevent data leakage. We randomly assign 256 sequences for training and 128 for testing. From these splits, we sample 2,200 training pairs and 1,100 testing pairs to form the final benchmark. This protocol ensures diverse motion patterns and scene structures while enabling reproducible quantitative evaluation. Detailed dataset statistics and qualitative examples are provided in Sec.~\ref{supp:db} of the supplementary document.

\vspace{-2mm}
\section{Proposed Method}

\begin{figure}[tb]
  \centering
  \includegraphics[width=\linewidth]{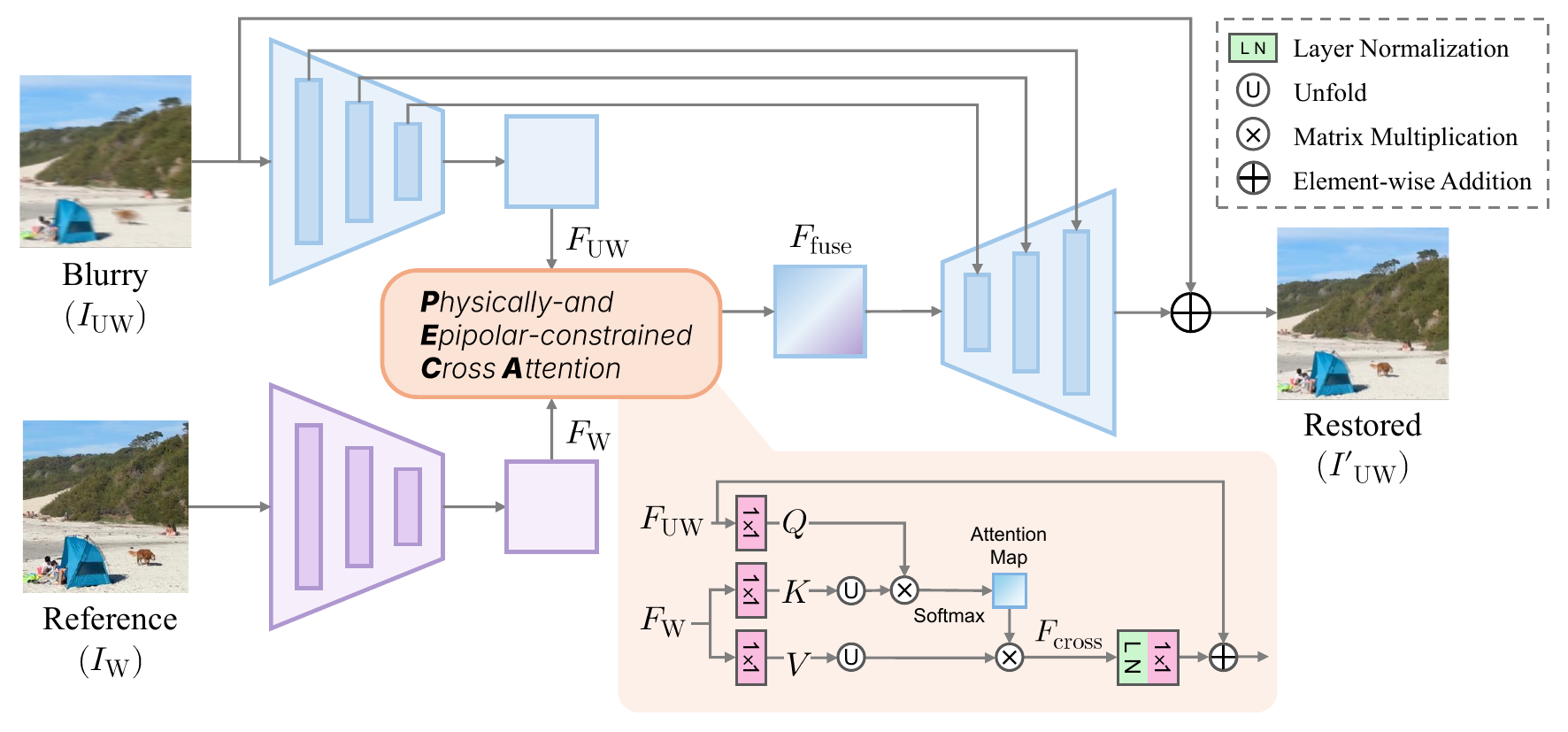}
  \vspace{-5mm}
  \caption{Overview of the proposed physically- and epipolar-constrained cross attention (PECA). The framework consists of dual encoders for the blurry ultra-wide input and the sharp wide reference, followed by the PECA module and residual feature fusion.}
  \label{fig:overview}
  \vspace{-3mm}
\end{figure}

\subsection{Design Principles}

We design PECA to address heterogeneous stereo deblurring in practical mobile capture, where blur is systematically asymmetric across views, and stereo pairs are provided through a device-level rectification pipeline, producing approximately epipolar-aligned observations.
PECA is guided by following three principles. 
\textbf{(i) Asymmetric restoration:} The wide view serves as a sharp reference while the ultra-wide view is blur-degraded; cross-view interaction should therefore be selective to enhance the target view without enforcing symmetric restoration. 
\textbf{(ii) Geometry- and physics-constrained correspondence:} Rectification restricts valid matches to epipolar scanlines, stereo geometry determines the disparity direction, and camera parameters provide a physically feasible disparity bound. PECA uses these priors to confine attention to a compact and feasible epipolar window. 
\textbf{(iii) Robust fusion without explicit matching or masks:} Dense matching or occlusion masks can be costly and sensitive to sensor heterogeneity and residual rectification errors. Instead, PECA relies on confidence-weighted attention within the constrained window and residual fusion to naturally attenuate unreliable regions.
This design produces an efficient and practical cross-view module that exploits rectified stereo geometry while remaining robust to partial visibility and device-level imperfections.

\vspace{-2mm}
\subsection{Overall Framework}
\vspace{-1mm}
The goal of our framework is to restore a sharp ultra-wide image $I'_\mathrm{UW}$ from a blurry ultra-wide observation $I_\mathrm{UW}$ by leveraging a synchronized sharp wide-view reference $I_\mathrm{W}$. 
In a heterogeneous stereo camera system, blur is typically concentrated in the ultra-wide stream, whereas the wide stream retains higher spatial fidelity.
We therefore formulate restoration asymmetrically, restoring $I_\mathrm{UW}$ while using $I_\mathrm{W}$ as a guidance source. 

As illustrated in Fig.~\ref{fig:overview}, we adopt a dual-branch encoder architecture.
The primary (top) branch encodes $I_\mathrm{UW}$ into self-view features $F_\mathrm{UW}$, while the reference (bottom) branch encodes $I_\mathrm{W}$ into reference features $F_\mathrm{W}$ that preserve sharp  structural cues.
The proposed PECA module retrieves cross-view features within a physically and geometrically constrained epipolar window, so the guidance signal is formed only from correspondence candidates that are plausible under rectified stereo geometry.

We handle partial visibility through confidence-modulated fusion. Some locations admit reliable cross-view correspondence, while others suffer from occlusions, texture ambiguity, or residual rectification errors. 
PECA captures this variation through its attention responses within the constrained window. When the attention weights concentrate on a few candidates, the module produces a strong cross-view guidance signal. When the weights spread out across the window, the guidance signal becomes weak and less informative. Residual fusion then injects the cross-view signal into the primary features while preserving the self-view pathway as an identity residual, enabling reference-guided enhancement in well-matched regions and relying on the self-view residual path in regions with weak or ambiguous correspondence.

Building on this overview, we next present PECA as a Transformer-style cross attention mechanism tailored to rectified stereo, where $F_\mathrm{UW}$ and $F_\mathrm{W}$ are mapped to query–key–value embeddings and attention is evaluated only within a physically bounded epipolar window to produce cross-view guidance features.

\begin{figure}[tb]
  \centering
  \includegraphics[width=\linewidth]{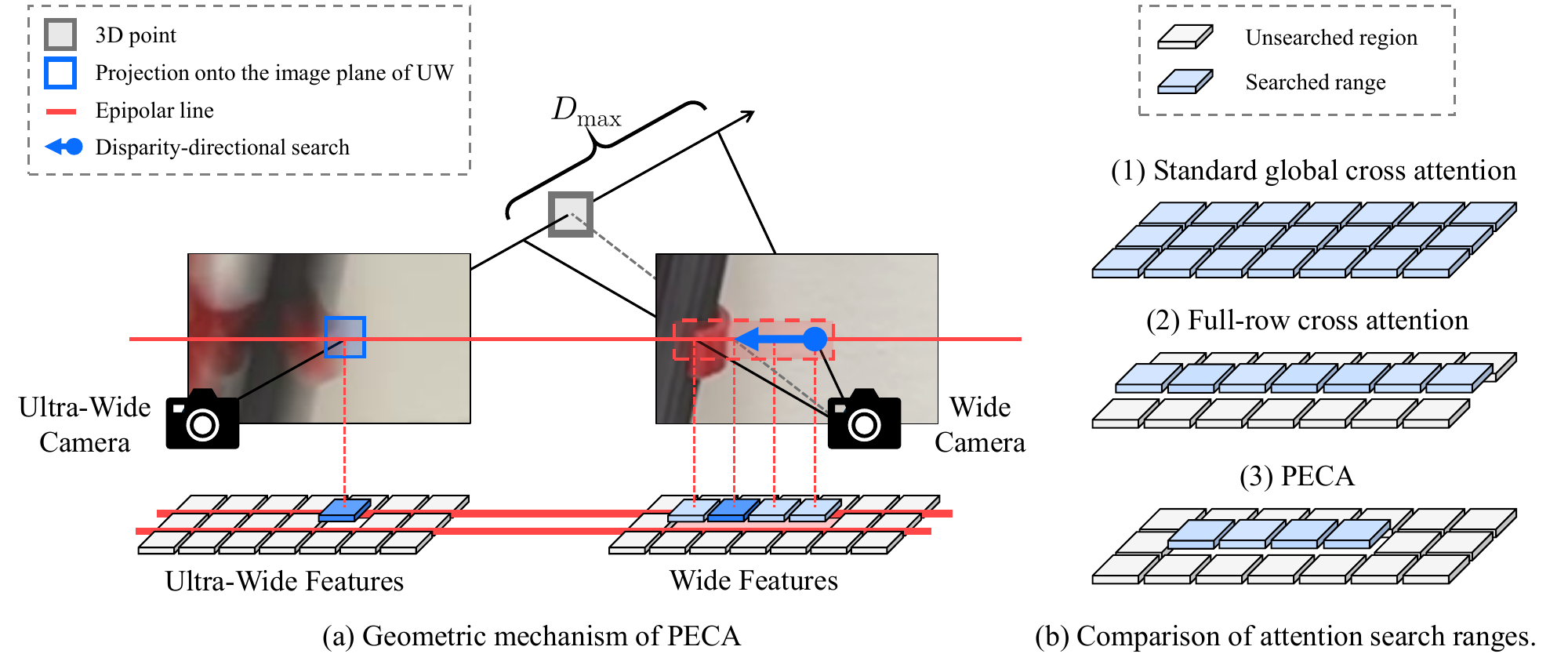}
  \vspace{-5mm}
  \caption{An illustration of the proposed PECA module. (a) The physically derived disparity bound $D_{\max}$ restricts correspondence search along the epipolar line. (b) Progressive restriction of attention search space: global cross attention (full 2D), full-row cross attention (full 1D scanline), and PECA (constrained 1D disparity window). }
  \label{fig:mechanism}
  \vspace{-3mm}
\end{figure}

\subsection{Physically- and epipolar-constrained cross attention (PECA)}
\label{sec4.3:peca}
\vspace{-1mm}
\subsubsection{Feature embeddings}
Let $F_\mathrm{UW} \in \mathbb{R}^{H \times W \times C}$ denote the latent feature from the primary (ultra-wide) branch and $F_\mathrm{W} \in \mathbb{R}^{H \times W \times C}$ denote the latent feature from the reference (wide) branch at the same spatial scale.
To perform cross attention, we map them into a shared embedding space via pointwise projections as follows:\begin{equation}
Q = \mathrm{Proj}(F_\mathrm{UW};W_q),\enspace
K = \mathrm{Proj}(F_\mathrm{W};W_k),\enspace\text{and}\enspace
V = \mathrm{Proj}(F_\mathrm{W};W_v),
\end{equation}
where $\mathrm{Proj}(\cdot;W)$ denotes a pointwise convolution parameterized by $W$, and $W_q$, $W_k$, and $W_v$ are the projection weights for query, key, value, respectively.
This design forms the queries from the degraded ultra-wide features, while extracting keys and values from the sharp wide reference, thereby enabling asymmetric cross-view retrieval.

\vspace{-2mm}
\subsubsection{Physically derived disparity constraint}
As illustrated in~\ref{fig:mechanism}(a), instead of searching across the entire image or full epipolar line, we use stereo geometry to obtain a physics-guided upper bound on disparity and restrict correspondence search to a plausible range. 
Let $d_\mathrm{min}^\mathrm{W}$ and $d_\mathrm{min}^\mathrm{UW}$ denote the minimum focus distances of the wide and ultra-wide cameras, respectively, and define the effective minimum acquisition distance as
\begin{equation}
d_\mathrm{eff} = \max(d_\mathrm{min}^\mathrm{W}, d_\mathrm{min}^\mathrm{UW}).
\end{equation}
Given the baseline distance $B$ between camera centers and the focal length $f$ (in pixel units), the maximum physically feasible disparity is approximately bounded by $f \cdot B /d_\mathrm{eff}$. When PECA operates at a reduced feature resolution, we can map this bound to the feature grid using a down-scaling factor $S$, which yields
\begin{equation}
D_\mathrm{phys} = \left\lceil \frac{f \cdot B}{d_\mathrm{eff}} \times \frac{1}{S}\right\rceil,
\end{equation}
where $S$ accounts for input resizing and architectural down-sampling.
This quantity provides a geometry-consistent ceiling on the correspondence search range in the feature space.
For our W/UW pair, the input-level disparity bound $f \cdot B/d_{\mathrm{eff}}$ is approximately $70$ pixels, which corresponds to $D_{\mathrm{phys}}\approx17$ when $S=4$. 

It is worth noting that $D_\mathrm{phys}$ reflects only the geometric scaling induced by $S$, with coarser feature grids yielding proportionally smaller bounds.
Since PECA matches encoded features, each feature embedding summarizes a spatial neighborhood whose extent depends on the encoder design, including network depth and kernel configuration, \ie, the effective receptive field. As the receptive field grows, fewer discrete offsets along the epipolar line can be sufficient to cover the physically feasible disparity range.
We therefore treat $D_\mathrm{phys}$ as an upper bound and choose the operational window size $D_{\max}$ enforcing $D_{\max} \leq D_\mathrm{phys}$.

\vspace{-2mm}
\subsubsection{Epipolar-constrained attention}
Since device-provided rectification aligns stereo pairs along epipolar scanlines, valid correspondences for a query location $(i,j)$ may lie on the same row $i$. 
Under the stereo-camera configuration, disparity is horizontal and bounded in a single direction along the scanline. Using the operational window size $D_{\max}$ chosen under the upper bound $D_\mathrm{phys}$, we define the localized 1D search window as
\begin{equation}
\Omega_{i,j} = \{ (i,k) \mid j - D_{\max} \le k \le j \}.
\end{equation}
The cross-view feature at $(i,j)$ is then computed by attending to the candidates in $\Omega_{i,j}$ as
\begin{equation}
F_\mathrm{cross}(i,j)
=
\sum_{(i,k) \in \Omega_{i,j}}
\operatorname{Softmax}_{k}
\left(
\frac{\mathcal{D}_{\cos}(Q_{i,j}, K_{i,k})}{\tau}
\right)\cdot
V_{i,k},
\label{eq:f_cross}
\end{equation}
where $\mathcal{D}_{\cos}(\cdot,\cdot)$ denotes the cosine similarity, and $\tau$ is the softmax temperature controlling the sharpness of the attention distribution.

In regions with reliable cross-view correspondence, attention weights concentrate around geometrically consistent matches, enabling effective transfer of sharp textures. 
In regions affected by occlusions, ambiguous textures, or residual rectification errors, the similarity scores become less distinctive within $\Omega_{i,j}$, producing diffuse weights and weaker guidance. 
This confidence-modulated response, together with residual fusion, allows PECA to emphasize reference information when correspondence is reliable and to rely on the self-view pathway otherwise, without explicit matching or occlusion masks.

\vspace{-2mm}
\subsubsection{Computational complexity}

As shown in Fig.~\ref{fig:mechanism}(b-1), standard global cross attention performs an unconstrained 2D search over the spatial domain, incurring $\mathcal{O}((HW)^2)$ complexity. 
By exploiting stereo geometry, prior stereo SR methods~\cite{wang2019learning, wang2021symmetric, chu2022nafssr} reduce this to $\mathcal{O}(HW^2)$ by restricting correspondence search to horizontal scanlines, which we refer to as full-row cross attention (Fig.~\ref{fig:mechanism}(b-2)).
PECA further constrains the search to a physically bounded 1D disparity window (Fig.~\ref{fig:mechanism}(b-3)), leading to a complexity of
\[
\mathcal{O}(HW D_{\max}),
\]
where $D_{\max} \ll W$ in practical mobile stereo configurations. 
This restriction significantly reduces computational overhead while preserving geometrically valid correspondence.

\vspace{-1mm}
\subsection{Complementary Feature Fusion}
\vspace{-1mm}
Finally, cross-view features are integrated with self-view features via residual summation:
\begin{equation}
F_\mathrm{fuse}
=
F_\mathrm{UW}
+
\mathrm{Proj}(F_\mathrm{cross};W_\mathrm{cross}),
\end{equation}
where $W_\mathrm{cross}$ is the learnable projection weights for cross-view feature aligning feature dimensions.
Since the magnitude of $F_\mathrm{cross}$ is governed by attention confidence within the physically constrained window, residual fusion effectively acts as a soft, data-adaptive gating mechanism. 
When confident matches exist, cross-view enhancement becomes dominant. 
Otherwise, when attention responses are weak or inconsistent, the identity pathway ensures stable self-deblurring.

Through this complementary interaction, PECA exploits cross-view redundancy while maintaining robustness under occlusion and geometric uncertainty, providing an efficient and physically grounded solution for heterogeneous stereo deblurring.

\vspace{-1mm}
\section{Experiments}
\vspace{-1mm}
\subsection{Experimental Setup}
\vspace{-1mm}
All models were implemented in PyTorch and trained on NVIDIA GPUs. 
We evaluated all methods on the proposed HSD dataset (see Sec.~\ref{hsd} for details).
To demonstrate its architectural generality, PECA was incorporated into three representative restoration backbones: CNN-based XYDeblur~\cite{ji2022xydeblur}, Transformer-based Restormer~\cite{zamir2022restormer}, and the simplified NAFNet~\cite{chen2022simple}.
Each backbone was trained for 400K iterations with a batch size of 8 using the AdamW~\cite{loshchilov2017decoupled} optimizer and a cosine annealing learning rate scheduler. 

Optimization hyperparameters follow the original settings of each backbone, with the initial learning rate adjusted for our batch size under the standard training heuristics~\cite{goyal2017accurate}.
Specifically, the learning rate, weight decay, and AdamW $\beta$ parameters were set to $1\times10^{-4}$, $1\times10^{-3}$, and $(0.9, 0.9)$ for XYDeblur, $3\times10^{-4}$, $1\times10^{-4}$, and $(0.9, 0.999)$ for Restormer, and $5\times10^{-4}$, $1\times10^{-3}$, and $(0.9, 0.9)$ for NAFNet, respectively.

We applied random cropping, gamma, and shot/read noise augmentation. Specifically, input images are randomly cropped into patches of size $128\times128$ for XYDeblur and Restormer, and $256\times256$ for NAFNet.
To ensure a fair comparison of PECA across architectures, we fixed the spatial resolution of the deepest latent features to $32\times32$ for all backbones, following the default configuration of XYDeblur.
Accordingly, NAFNet was configured as a three-level architecture with [1, 1, 28] and [1, 1, 1] NAFBlocks in the encoder and decoder, respectively. For Restormer, the downsampling operation preceding the fourth encoder stage was removed.

Note that computational complexity (MACs) was measured at an input resolution of $720\times1280$.
Unless otherwise specified, we fix the softmax temperature to $\tau=0.01$ and use a single default window size $D_{\max}=5$ across all backbones to avoid backbone-specific tuning.
Sensitivity analyses are provided in Sec.~\ref{exp:params}, with additional qualitative results in Supplementary Sec.~\ref{supp:hyper}.

\begin{table}[t]
\centering
\caption{Performance and complexity comparison on HSD. Four model variants are evaluated across three backbones representing distinct architectural paradigms.}
\vspace{-2mm}
\label{tab:sota}
{\setlength{\tabcolsep}{6.5pt}
\scalebox{0.75}{
\begin{tabular}{lllcccc}
\toprule
\textbf{Backbone} & \textbf{Variant} & \textbf{Input} & \textbf{PSNR$\uparrow$} & \textbf{SSIM$\uparrow$} & \textbf{Params. (M)$\downarrow$}  & \textbf{MACs (G) $\downarrow$} \\
\midrule
XYDeblur~\cite{ji2022xydeblur} & baseline & Single & 30.76 & 0.9444 & 4.92 & 1056.5 \\
 & + channel-expansion & Single & 30.92 & 0.9462 & 7.68 & 1650.1 \\
 & + stereo concatenation & Stereo & 31.96 & 0.9595 & 7.37 & 1373.5 \\
 & + \textbf{PECA (Ours)} & Stereo & \textbf{32.22} & \textbf{0.9620} & 7.42 & 1376.4 \\

\midrule
Restormer~\cite{zamir2022restormer} & baseline & Single & 30.94 & 0.9444 & 12.59 & 2139.8 \\
 & + channel-expansion & Single & 31.17 & 0.9471 & 22.17 & 3754.7 \\
 & + stereo concatenation & Stereo & 31.83 & 0.9581 & 19.94 & 2851.3 \\
 & + \textbf{PECA (Ours)} & Stereo & \textbf{32.47} & \textbf{0.9636} & 20.05 & 2857.8 \\
\midrule
NAFNet~\cite{chen2022simple} & baseline & Single & 32.16 & 0.9579 & 17.08 & 832.6 \\
 & + channel-expansion & Single & 32.28 & 0.9590 & 38.22 & 1860.5 \\
 & + stereo concatenation & Stereo & 32.24 & 0.9604 & 33.38 & 1567.4 \\
 & + \textbf{PECA (Ours)} & Stereo & \textbf{32.92} & \textbf{0.9669} & 34.17 & 1589.6 \\
\bottomrule
\end{tabular}}}
\vspace{-3mm}
\end{table}

\vspace{-2mm}
\subsection{Comparison on the HSD Dataset}
\vspace{-1mm}

We evaluate PECA on three representative restoration backbones, XYDeblur~\cite{ji2022xydeblur}, Restormer~\cite{zamir2022restormer}, and NAFNet~\cite{chen2022simple}. 
For each backbone, we compare four variants: the single-image baseline, a channel-expansion model that increases capacity in the single-view setting, a naive stereo concatenation model, and the proposed PECA module. 
Parameter counts are kept comparable whenever possible for fair comparison.

\begin{figure}[tb]
  \centering
  \includegraphics[width=\linewidth]{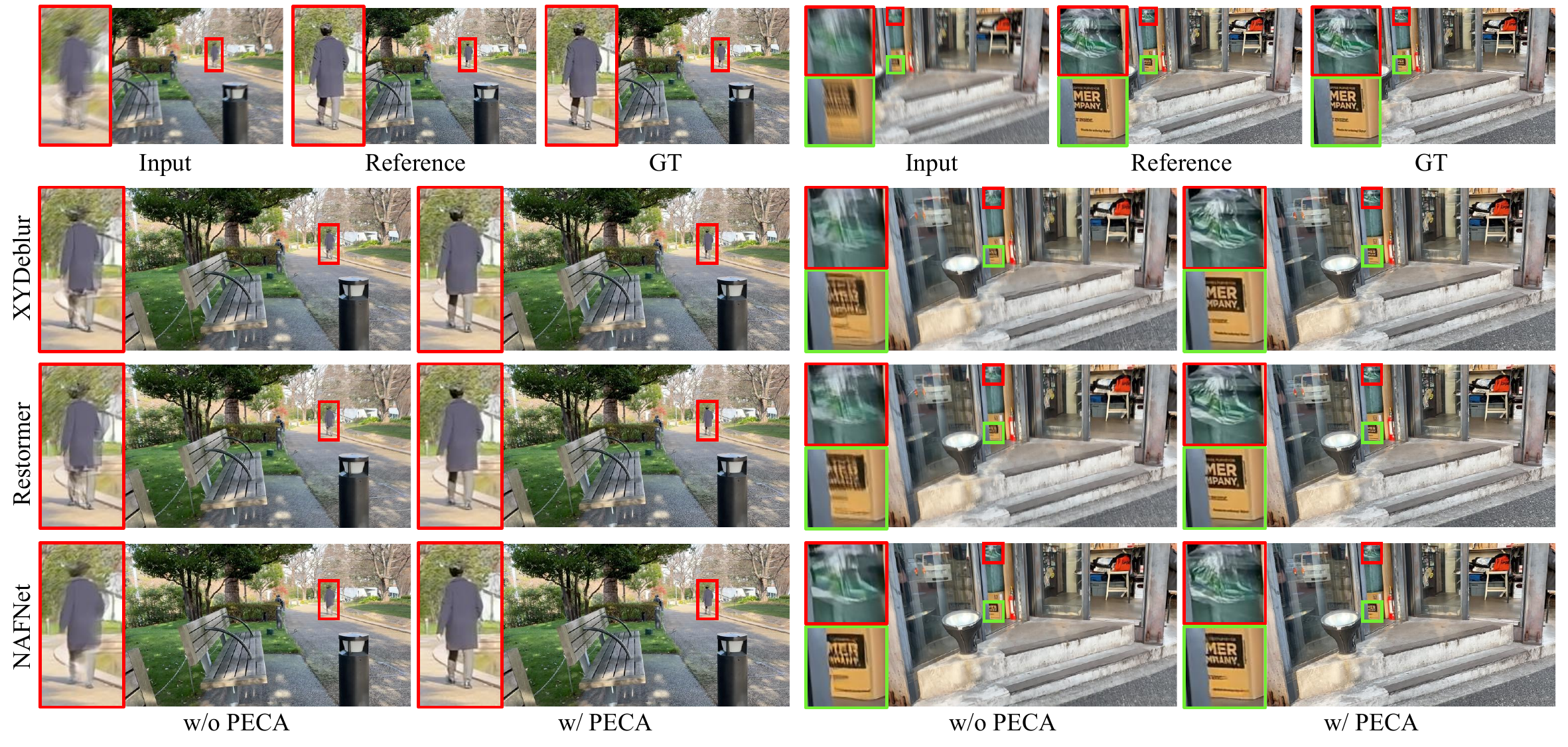}
  \vspace{-5mm}
  \caption{Qualitative results on the HSD dataset for three backbones, evaluated based on the presence of PECA. Three zoomed regions focusing on large-scale blur, transparent vinyl, and fine-grained text are highlighted.}
  \label{fig:test}
  \vspace{-3mm}
\end{figure}

Table~\ref{tab:sota} reports quantitative results on the HSD dataset. 
Incorporating the synchronized wide-view reference consistently improves performance over single-image baselines across all backbones.
The gains from PECA are not simply explained by increased model capacity. Despite using more parameters and higher computational cost, the channel-expansion models underperform PECA.
For example, with Restormer, PECA improves PSNR by +1.30\,dB over channel-expansion while requiring lower computational cost.
This suggests that the improvement of PECA stems from structured cross-view feature retrieval rather than simply scaling up the network.
Even naive stereo concatenation yields noticeable gains (\eg, +1.20\,dB for XYDeblur), confirming that cross-view information is beneficial for asymmetric deblurring. 
Compared with stereo concatenation, PECA consistently achieves higher restoration quality (\eg, +0.68\,dB on NAFNet), indicating that explicitly modeling cross-view correspondence is critical beyond simple feature fusion.
Additional analysis in Supplementary Sec.~\ref{supp:occ} examines this behavior under severe occlusion, where PECA maintains positive gains while naive stereo concatenation can fall below the single-view baseline.
Qualitative results in Fig.~\ref{fig:test} further show that PECA restores sharper edges and fine textures, particularly in severely blurred regions and small text patterns, while avoiding over-smoothed reconstructions.
Overall, the consistent gains across CNN-, Transformer-, and NAFNet-based architectures demonstrate that PECA provides a robust and backbone-agnostic mechanism for heterogeneous stereo deblurring.

\begin{figure}[t]
\centering
\includegraphics[width=\linewidth]{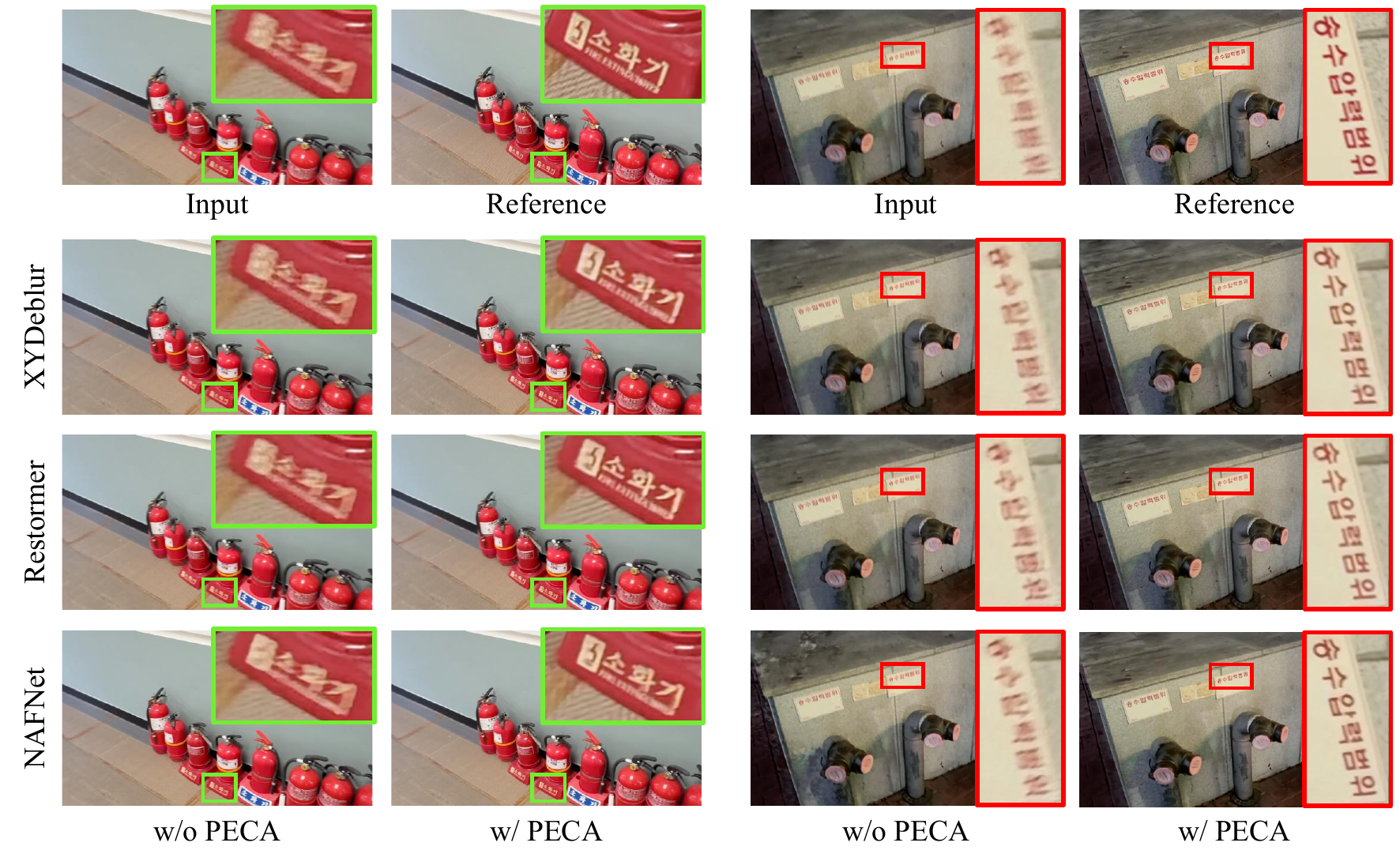}
\vspace{-5mm}
\caption{Qualitative comparison on real handheld stereo captures without ground truth.}
\label{fig:real}
\end{figure}

\vspace{-2mm}
\subsection{Real-world Qualitative Evaluation}
\vspace{-1mm}

Beyond the temporally integrated pairs in HSD, we additionally provide qualitative results on real handheld heterogeneous stereo captures, where the target blur arises directly from natural capture dynamics rather than the dataset construction protocol.
These sequences contain severe motion blur with depth variation and frequent occlusions, and no ground-truth sharp images are available

Fig.~\ref{fig:real} compares each baseline backbone with its PECA-enhanced counterparts. 
The baseline models often produce over-smoothed textures or unstable edge reconstruction under strong blur. 
In contrast, the PECA exhibits sharper structures and clearer fine details across all three architectures, indicating that the proposed reference-guided fusion remains effective under real handheld blur.
Although quantitative evaluation is not available for these real samples, the consistent qualitative improvements complement the paired benchmark results and support the practical applicability of PECA in realistic heterogeneous smartphone imaging conditions.

\vspace{-2mm}
\subsection{Ablation Study}
\vspace{-1mm}
\subsubsection{Effect of attention search ranges}
Table~\ref{tab:search_range} compares three cross-view attention designs that differ only in their correspondence search space. 
Global cross attention performs exhaustive 2D matching ($H×W$), which incurs high computational cost (852.2G MACs for the attention module).
Under device-provided rectification, valid correspondences are strongly concentrated on the epipolar scanline; hence, if similarity matching is effective, the off-row candidates in global search are unlikely to receive high attention weights, making global and full-row-based search yield similar restoration behavior in practice.
Full-row attention explicitly restricts the search to the epipolar line, reducing the number of MACs by $112\times$, but it still considers the range along the scanline. In asymmetric deblurring, blur suppresses discriminative high-frequency cues and increases pattern repetition along a row, so many distant locations become competitive, leading to diffuse attention and limited gains (+0.03\,dB) over the global attention model.

In contrast, PECA further constrains matching to a geometrically plausible disparity window, filtering out scanline candidates that are physically unlikely to correspond to the query. This reduces ambiguity and computation costs, resulting in a PSNR of 32.22\,dB (+1.50\,dB/+1.47\,dB over the global/full-row models) with only 2.9G MACs for the attention module.
Additional qualitative comparisons and the top-3 attention distribution analysis corresponding to these results are provided in the supplementary document, in Sec.~\ref{supp:search_range} and Sec.~\ref{supp:rel_pos}, respectively.

\begin{table}[t]
\centering
\caption{Effect of search-space design on deblurring performance. 
All attention variants use identical $1\times1$ projection layers and cosine similarity with $\tau=0.01$.}
\vspace{-3mm}
\label{tab:search_range}
{\setlength{\tabcolsep}{6.5pt}
\scalebox{0.8}{
\begin{tabular}{lccccc}
\toprule
\textbf{Method} & \textbf{Search space}  &\textbf{PSNR $\uparrow$} & \textbf{SSIM $\uparrow$} & \textbf{Module MACs (G) $\downarrow$} & \textbf{Total MACs (G) $\downarrow$}\\ \midrule

Global& 2D ($H×W$) & 30.72 & 0.9439 & 852.2 & 2225.7 \\
Full-row& 1D ($W$)  & 30.75 & 0.9438 & 7.6 & 1381.0\\ 
PECA& 1D ($D_{\max}$) & \textbf{32.22} & \textbf{0.9620} & \textbf{2.9} & \textbf{1376.4} \\

\bottomrule
\end{tabular}}}
\vspace{-3mm}
\end{table}

\subsubsection{Parameter sensitivity of $D_{\max}$ and $\tau$}
\label{exp:params}
\begin{figure}[tb]
  \centering
  \includegraphics[width=\linewidth]{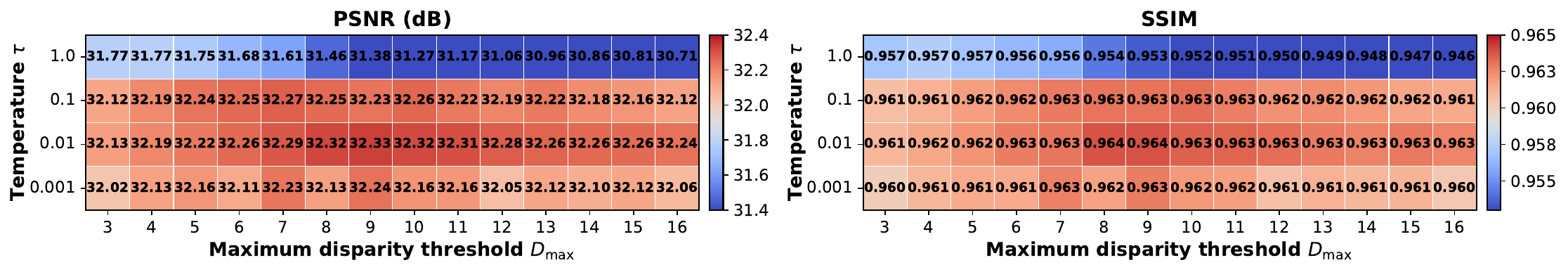}
    \vspace{-5mm}
  \caption{PSNR and SSIM heatmaps over $D_{\max}$ and $\tau$ with the XYDeblur backbone.
  }
  \label{fig:params}
  \vspace{-5mm}
\end{figure}

Fig.~\ref{fig:params} analyzes how the epipolar window size $D_{\max}$ and the softmax temperature $\tau$ interact in PECA using the XYDeblur~\cite{ji2022xydeblur} backbone.
When the attention distribution is sufficiently sharp (\eg, $\tau{=}0.01$), PECA achieves consistently strong performance across a range of window sizes, from $D_{\max}{=}3$ up to $D_{\max}{=}16$ (which is close to the physically feasible disparity range $D_\mathrm{phys}$). We observe the best PSNR at $D_{\max}{=}9$ for XYDeblur when $\tau{=}0.01$, while smaller windows deliver comparable accuracy with reduced attention cost. 
As discussed in Sec.~\ref{sec4.3:peca}, the preferred $D_{\max}$ can vary across backbones due to differences in the effective receptive field, which depends on kernel configuration and network depth. Since our goal is to verify the general improvement of PECA rather than to tune $D_{\max}$ for each specific architecture, unless otherwise specified, we use a single default window size $D_{\max}{=}5$ in our main experiments. 

Regarding $\tau$, a small temperature yields a peaky attention distribution that better isolates reliable correspondences, whereas a large temperature (\eg, $\tau{=}1.0$) produces relatively diffuse weights due to the bounded cosine similarity, which may weaken discrimination and degrades restoration quality. In this diffuse regime, expanding $D_{\max}$ can be less beneficial since a wider window mainly introduces additional ambiguous candidates along the scanline.

\vspace{-2mm}
\section{Conclusion}
\vspace{-2mm}
In this paper, we addressed asymmetric stereo deblurring in heterogeneous stereo camera systems, where hardware-induced quality discrepancies naturally arise between views.
We proposed PECA, which embeds physical geometric priors into cross-view attention. This approach enables efficient and reliable information transfer by replacing unconstrained global search with localized and directional epipolar-aware matching.
We also introduced the HSD Dataset, a real-world benchmark designed to evaluate asymmetric stereo deblurring in heterogeneous stereo camera systems.
Extensive experiments demonstrate that PECA consistently improves deblurring performance across diverse network architectures while maintaining computational efficiency suitable for practical deployment.
These results validate the effectiveness of geometry-constrained cross-view modeling in addressing asymmetric degradation under practical heterogeneous camera settings.
We believe that incorporating physically grounded constraints into cross-view attention provides a principled foundation for future asymmetric multi-camera restoration methods.

\clearpage

%
%
\bibliographystyle{splncs04}
\bibliography{main}

@String(AAAI  = {AAAI})

@inproceedings{ji2022xydeblur,
  title={XYDeblur: Divide and conquer for single image deblurring},
  author={Ji, Seo-Won and Lee, Jeongmin and Kim, Seung-Wook and Hong, Jun-Pyo and Baek, Seung-Jin and Jung, Seung-Won and Ko, Sung-Jea},
  booktitle={Proceedings of the IEEE/CVF conference on computer vision and pattern recognition},
  pages={17421--17430},
  year={2022}
}

@inproceedings{chen2022simple,
  title={Simple baselines for image restoration},
  author={Chen, Liangyu and Chu, Xiaojie and Zhang, Xiangyu and Sun, Jian},
  booktitle={European conference on computer vision},
  pages={17--33},
  year={2022},
  organization={Springer}
}

@inproceedings{chu2022nafssr,
  title={Nafssr: Stereo image super-resolution using nafnet},
  author={Chu, Xiaojie and Chen, Liangyu and Yu, Wenqing},
  booktitle={Proceedings of the IEEE/CVF conference on computer vision and pattern recognition},
  pages={1239--1248},
  year={2022}
}

@inproceedings{cho2021rethinking,
  title={Rethinking coarse-to-fine approach in single image deblurring},
  author={Cho, Sung-Jin and Ji, Seo-Won and Hong, Jun-Pyo and Jung, Seung-Won and Ko, Sung-Jea},
  booktitle={Proceedings of the IEEE/CVF international conference on computer vision},
  pages={4641--4650},
  year={2021}
}

@inproceedings{liu2023real,
  title={Real-world image deblurring via unsupervised domain adaptation},
  author={Liu, Hanzhou and Li, Binghan and Lu, Mi and Wu, Yucheng},
  booktitle={International Symposium on Visual Computing},
  pages={148--159},
  year={2023},
  organization={Springer}
}

@inproceedings{zamir2022restormer,
  title={Restormer: Efficient transformer for high-resolution image restoration},
  author={Zamir, Syed Waqas and Arora, Aditya and Khan, Salman and Hayat, Munawar and Khan, Fahad Shahbaz and Yang, Ming-Hsuan},
  booktitle={Proceedings of the IEEE/CVF conference on computer vision and pattern recognition},
  pages={5728--5739},
  year={2022}
}

@inproceedings{kong2023efficient,
  title={Efficient frequency domain-based transformers for high-quality image deblurring},
  author={Kong, Lingshun and Dong, Jiangxin and Ge, Jianjun and Li, Mingqiang and Pan, Jinshan},
  booktitle={Proceedings of the IEEE/CVF conference on computer vision and pattern recognition},
  pages={5886--5895},
  year={2023}
}

@inproceedings{liu2025xyscannet,
  title={XYScanNet: A State Space Model for Single Image Deblurring},
  author={Liu, Hanzhou and Liu, Chengkai and Xu, Jiacong and Jiang, Peng and Lu, Mi},
  booktitle={Proceedings of the IEEE/CVF Conference on Computer Vision and Pattern Recognition},
  pages={779--789},
  year={2025}
}

@article{zhang2025survey,
  title={A Survey of Single Image Blind Motion Deblurring from Traditional to Deep Learning},
  author={Zhang, Tingting and Lu, Jiawei and Jin, Qiyu and Zeng, Tieyong},
  journal={ACM Computing Surveys},
  year={2025},
  publisher={ACM New York, NY}
}

@inproceedings{zou2023cross,
  title={Cross-view hierarchy network for stereo image super-resolution},
  author={Zou, Wenbin and Gao, Hongxia and Chen, Liang and Zhang, Yunchen and Jiang, Mingchao and Yu, Zhongxin and Tan, Ming},
  booktitle={Proceedings of the IEEE/CVF conference on computer vision and pattern recognition},
  pages={1396--1405},
  year={2023}
}

@inproceedings{sellent2016stereo,
  title={Stereo video deblurring},
  author={Sellent, Anita and Rother, Carsten and Roth, Stefan},
  booktitle={European conference on computer vision},
  pages={558--575},
  year={2016},
  organization={Springer}
}

@inproceedings{pan2017simultaneous,
  title={Simultaneous stereo video deblurring and scene flow estimation},
  author={Pan, Liyuan and Dai, Yuchao and Liu, Miaomiao and Porikli, Fatih},
  booktitle={Proceedings of the IEEE conference on computer vision and pattern recognition},
  pages={4382--4391},
  year={2017}
}

@article{pan2019joint,
  title={Joint stereo video deblurring, scene flow estimation and moving object segmentation},
  author={Pan, Liyuan and Dai, Yuchao and Liu, Miaomiao and Porikli, Fatih and Pan, Quan},
  journal={IEEE Transactions on Image Processing},
  volume={29},
  pages={1748--1761},
  year={2019},
  publisher={IEEE}
}

@inproceedings{zhou2019davanet,
  title={Davanet: Stereo deblurring with view aggregation},
  author={Zhou, Shangchen and Zhang, Jiawei and Zuo, Wangmeng and Xie, Haozhe and Pan, Jinshan and Ren, Jimmy S},
  booktitle={Proceedings of the IEEE/CVF Conference on Computer Vision and Pattern Recognition},
  pages={10996--11005},
  year={2019}
}

@inproceedings{wang2021dual,
  title={Dual-camera super-resolution with aligned attention modules},
  author={Wang, Tengfei and Xie, Jiaxin and Sun, Wenxiu and Yan, Qiong and Chen, Qifeng},
  booktitle={Proceedings of the IEEE/CVF international conference on computer vision},
  pages={2001--2010},
  year={2021}
}

@inproceedings{yue2024kedusr,
  title={Kedusr: Real-world dual-lens super-resolution via kernel-free matching},
  author={Yue, Huanjing and Cui, Zifan and Li, Kun and Yang, Jingyu},
  booktitle={Proceedings of the AAAI Conference on Artificial Intelligence},
  volume={38},
  number={7},
  pages={6881--6889},
  year={2024}
}

@inproceedings{kang2025dual,
  title={Dual-Lens Super-Resolution with Semantic-Enhanced Feature Matching and Adaptive Texture Transfer},
  author={Kang, Jinshi and Yao, Rui and Zhu, Hancheng and Sun, Kunyang and Li, Xixi and Zhao, Jiaqi and Zhou, Yong},
  booktitle={Journal of Physics: Conference Series},
  volume={3108},
  number={1},
  pages={012021},
  year={2025},
  organization={IOP Publishing}
}

@inproceedings{lee2022reference,
  title={Reference-based video super-resolution using multi-camera video triplets},
  author={Lee, Junyong and Lee, Myeonghee and Cho, Sunghyun and Lee, Seungyong},
  booktitle={Proceedings of the IEEE/CVF conference on computer vision and pattern recognition},
  pages={17824--17833},
  year={2022}
}

@inproceedings{kim2023efficient,
  title={Efficient reference-based video super-resolution (ervsr): Single reference image is all you need},
  author={Kim, Youngrae and Lim, Jinsu and Cho, Hoonhee and Lee, Minji and Lee, Dongman and Yoon, Kuk-Jin and Choi, Ho-Jin},
  booktitle={Proceedings of the IEEE/CVF Winter Conference on Applications of Computer Vision},
  pages={1828--1837},
  year={2023}
}

@inproceedings{zou2025refvsr++,
  title={RefVSR++: Exploiting Reference Inputs for Reference-based Video Super-resolution},
  author={Zou, Han and Suganuma, Masanori and Okatani, Takayuki},
  booktitle={2025 IEEE/CVF Winter Conference on Applications of Computer Vision (WACV)},
  pages={2756--2765},
  year={2025},
  organization={IEEE}
}

@inproceedings{xiaoasymmetric,
  title={Asymmetric Dual-Lens Video Deblurring},
  author={Xiao, Zeyu and Wang, Xinchao},
  booktitle={The Thirty-ninth Annual Conference on Neural Information Processing Systems},
  year={2025}
}

@inproceedings{rim2024deep,
  title={Deep hybrid camera deblurring for smartphone cameras},
  author={Rim, Jaesung and Lee, Junyong and Yang, Heemin and Cho, Sunghyun},
  booktitle={ACM SIGGRAPH 2024 Conference Papers},
  pages={1--11},
  year={2024}
}

@article{shekarforoush2023dual,
  title={Dual-camera joint deblurring-denoising},
  author={Shekarforoush, Shayan and Walia, Amanpreet and Brubaker, Marcus A and Derpanis, Konstantinos G and Levinshtein, Alex},
  journal={arXiv preprint arXiv:2309.08826},
  year={2023}
}

@article{lin2025learning,
  title={Learning parallax for stereo event-based motion deblurring},
  author={Lin, Mingyuan and Zhang, Chi and He, Chu and Yu, Lei},
  journal={IEEE Transactions on Circuits and Systems for Video Technology},
  year={2025},
  publisher={IEEE}
}

@inproceedings{wang2019learning,
  title={Learning parallax attention for stereo image super-resolution},
  author={Wang, Longguang and Wang, Yingqian and Liang, Zhengfa and Lin, Zaiping and Yang, Jungang and An, Wei and Guo, Yulan},
  booktitle={Proceedings of the IEEE/CVF conference on computer vision and pattern recognition},
  pages={12250--12259},
  year={2019}
}

@inproceedings{wang2021symmetric,
  title={Symmetric parallax attention for stereo image super-resolution},
  author={Wang, Yingqian and Ying, Xinyi and Wang, Longguang and Yang, Jungang and An, Wei and Guo, Yulan},
  booktitle={Proceedings of the IEEE/CVF Conference on Computer Vision and Pattern Recognition},
  pages={766--775},
  year={2021}
}

@inproceedings{nah2017deep,
  title={Deep multi-scale convolutional neural network for dynamic scene deblurring},
  author={Nah, Seungjun and Hyun Kim, Tae and Mu Lee, Kyoung},
  booktitle={Proceedings of the IEEE conference on computer vision and pattern recognition},
  pages={3883--3891},
  year={2017}
}

@inproceedings{rim2020real,
  title={Real-world blur dataset for learning and benchmarking deblurring algorithms},
  author={Rim, Jaesung and Lee, Haeyun and Won, Jucheol and Cho, Sunghyun},
  booktitle={European conference on computer vision},
  pages={184--201},
  year={2020},
  organization={Springer}
}

@inproceedings{nah2019ntire,
  title={Ntire 2019 challenge on video deblurring and super-resolution: Dataset and study},
  author={Nah, Seungjun and Baik, Sungyong and Hong, Seokil and Moon, Gyeongsik and Son, Sanghyun and Timofte, Radu and Mu Lee, Kyoung},
  booktitle={Proceedings of the IEEE/CVF conference on computer vision and pattern recognition workshops},
  pages={0--0},
  year={2019}
}

@inproceedings{shen2019human,
  title={Human-aware motion deblurring},
  author={Shen, Ziyi and Wang, Wenguan and Lu, Xiankai and Shen, Jianbing and Ling, Haibin and Xu, Tingfa and Shao, Ling},
  booktitle={Proceedings of the IEEE/CVF international conference on computer vision},
  pages={5572--5581},
  year={2019}
}

@article{delbracio2021mobile,
  title={Mobile computational photography: A tour},
  author={Delbracio, Mauricio and Kelly, Damien and Brown, Michael S and Milanfar, Peyman},
  journal={Annual review of vision science},
  volume={7},
  number={1},
  pages={571--604},
  year={2021},
  publisher={Annual Reviews}
}

@article{kim2021generation,
  title={Generation of stereo images from the heterogeneous cameras},
  author={Kim, SeongKi},
  journal={Journal homepage: http://iieta. org/journals/i2m},
  volume={20},
  number={2},
  pages={73--78},
  year={2021}
}

@article{loshchilov2017decoupled,
  title={Decoupled weight decay regularization},
  author={Loshchilov, Ilya and Hutter, Frank},
  journal={arXiv preprint arXiv:1711.05101},
  year={2017}
}

@article{goyal2017accurate,
  title={Accurate, large minibatch sgd: Training imagenet in 1 hour},
  author={Goyal, Priya and Doll{\'a}r, Piotr and Girshick, Ross and Noordhuis, Pieter and Wesolowski, Lukasz and Kyrola, Aapo and Tulloch, Andrew and Jia, Yangqing and He, Kaiming},
  journal={arXiv preprint arXiv:1706.02677},
  year={2017}
}

@inproceedings{min2025s2m2,
  title={S2M2: Scalable Stereo Matching Model for Reliable Depth Estimation},
  author={Min, Junhong and Jeon, Youngpil and Kim, Jimin and Choi, Minyong},
  booktitle={Proceedings of the IEEE/CVF International Conference on Computer Vision},
  pages={26729--26739},
  year={2025}
}

@article{zhang2024stereo,
  title={Stereo image restoration via attention-guided correspondence learning},
  author={Zhang, Shengping and Yu, Wei and Jiang, Feng and Nie, Liqiang and Yao, Hongxun and Huang, Qingming and Tao, Dacheng},
  journal={IEEE Transactions on Pattern Analysis and Machine Intelligence},
  volume={46},
  number={7},
  pages={4850--4865},
  year={2024},
  publisher={IEEE}
}

\appendix
\onecolumn
\newpage

\renewcommand\thesection{\Alph{section}}
\renewcommand\thetable{\Alph{table}}
\renewcommand\thefigure{\Alph{figure}}
\newcommand\numberthis{\addtocounter{equation}{1}\tag{\theequation}}

\begin{center}
    \Large \textbf{A Benchmark for Heterogeneous Stereo Deblurring with PECA}\par
    \Large \textbf{\textit{--Supplementary Document--}}
\end{center}



\setcounter{table}{0}
\setcounter{figure}{0}
\setcounter{theorem}{0}
\setcounter{page}{1}
\setlength{\textfloatsep}{15pt}

\begin{figure}[h]
\centering
\includegraphics[width=\linewidth]{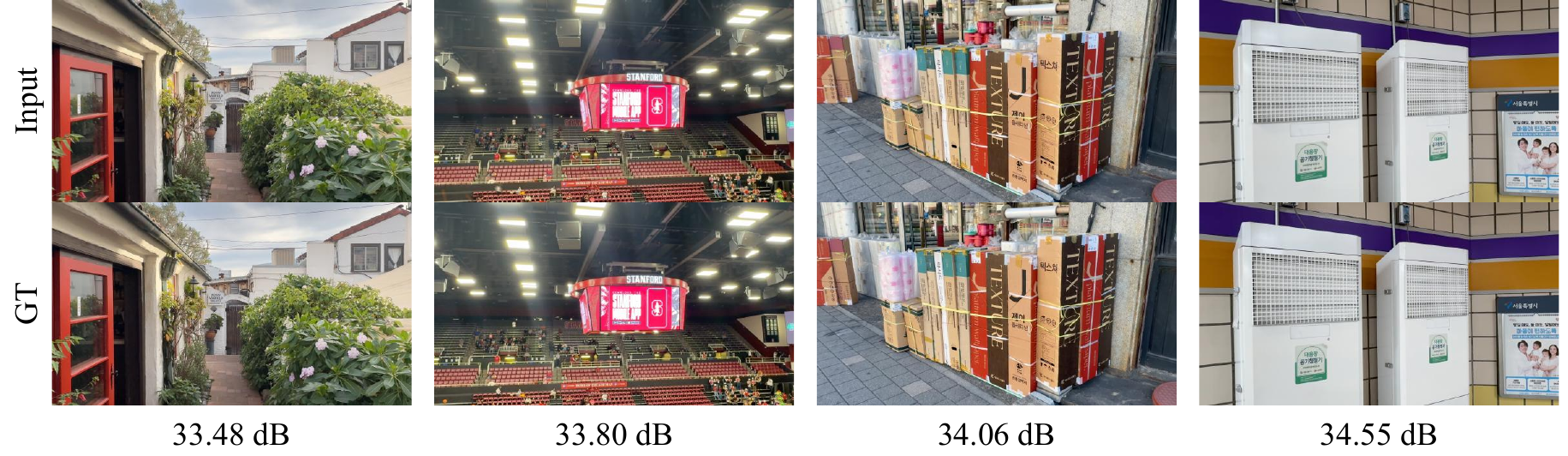}
\caption{Examples of image pairs with PSNR above 33 dB.}
\label{fig:static}
\vspace{-8mm}
\end{figure}

\section{Dataset Statistics and Qualitative Examples}
\label{supp:db}

\begin{figure}[b]
\centering
\begin{subfigure}[t]{0.48\linewidth}
\centering
\includegraphics[width=\linewidth]{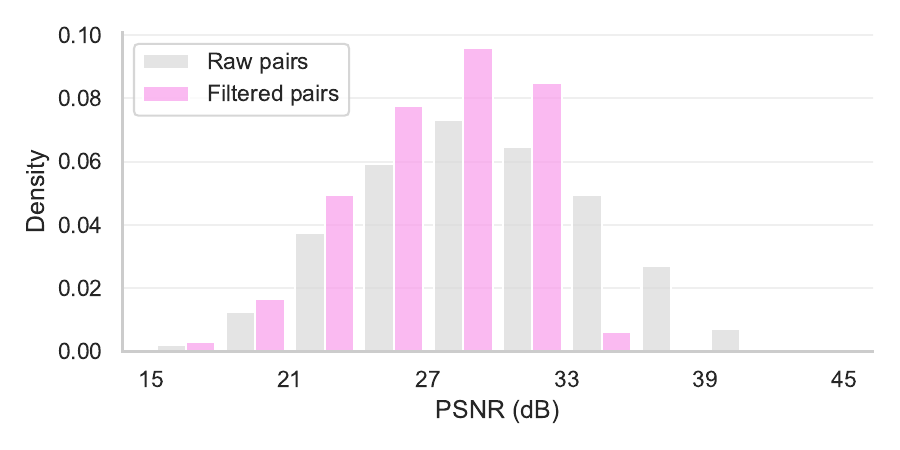}
\caption{Before and after filtering.}
\end{subfigure}
\hfill
\begin{subfigure}[t]{0.48\linewidth}
\centering
\includegraphics[width=\linewidth]{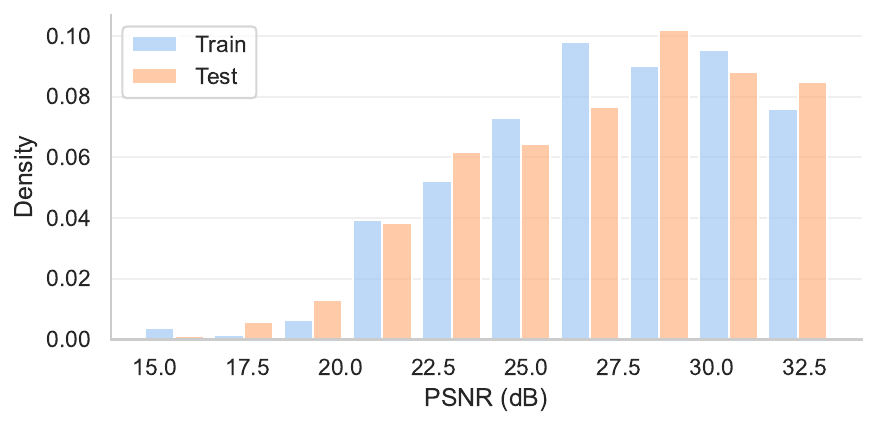}
\caption{Training and test splits.}
\end{subfigure}
\caption{Distribution of PSNR in the HSD dataset.}
\label{fig:hist}
\end{figure}

\begin{figure}[t]
\centering
\includegraphics[width=0.9\linewidth]{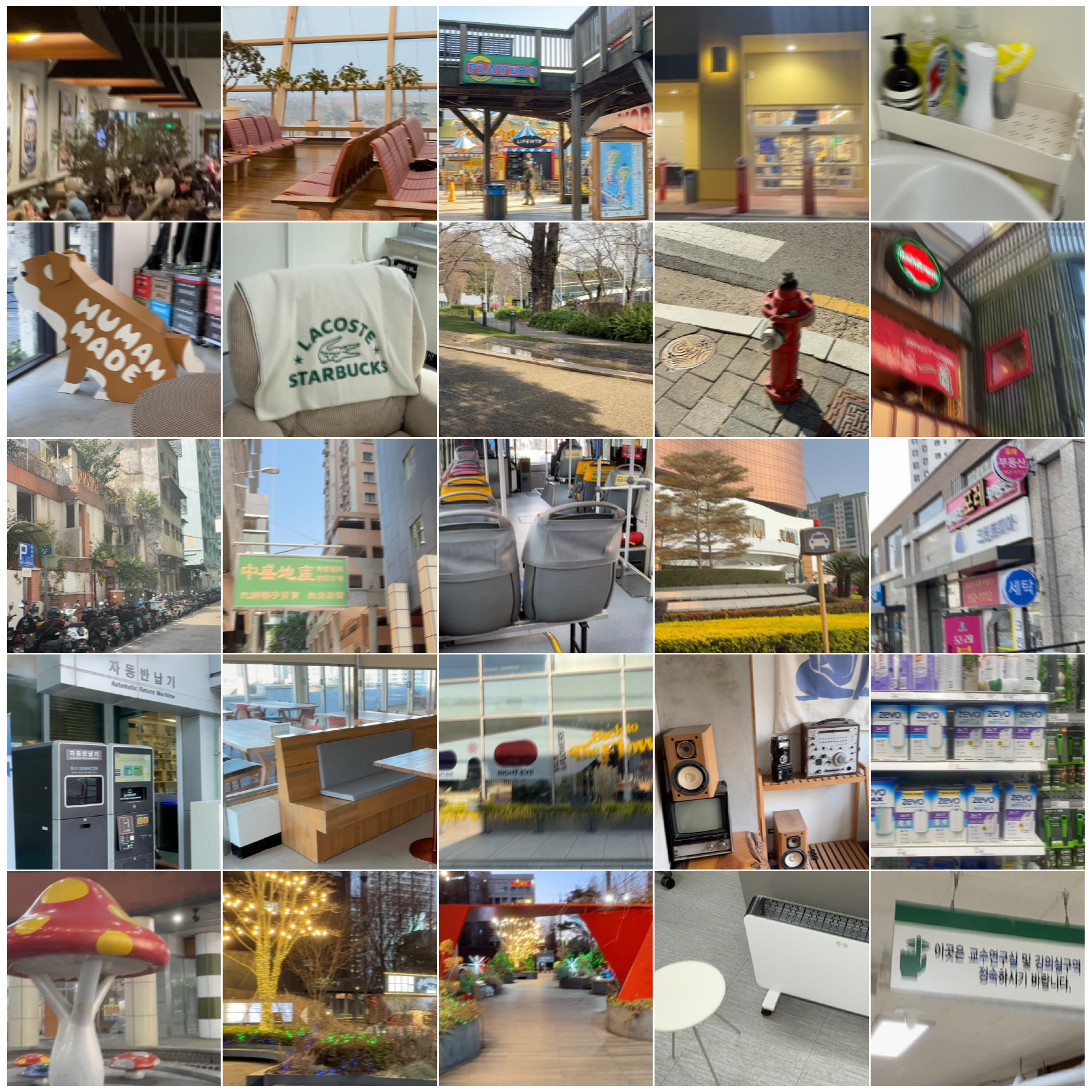}
\caption{Representative blurry image patches from the HSD dataset, illustrating diverse scenarios such as indoor/outdoor scenes, varying lighting and blur levels, and different object densities and depth structures.}
\label{fig:div}
\end{figure}

To analyze dataset characteristics and identify nearly static scenes, we compute the PSNR between blurry and sharp ultra-wide image pairs.
Through visual inspection, we observe that pairs with PSNR above approximately 33 dB correspond to scenes with negligible motion blur.
Since such samples provide insufficient supervision for motion deblurring, we exclude pairs whose PSNR exceeds this threshold.
Examples of such nearly static scenes are shown in Fig.~\ref{fig:static}.

After filtering, we obtain 37,623 image pairs from 384 sequences.
We randomly split the dataset into training and test sets at the sequence level to avoid data leakage between the two splits. 
Specifically, we use 256 sequences for training and 128 sequences for testing. 
We then randomly sample 2,200 and 1,100 image pairs from the training and test sequences, respectively.
Figs.~\ref{fig:hist}(a) and (b) illustrate the PSNR distributions before and after filtering, and those of the training and test splits, respectively. The two splits exhibit similar distributions, suggesting that the partitioning does not introduce significant statistical bias.

The HSD dataset covers a wide variety of scenes captured across multiple cities, including Seoul, Busan, Macau, Tokyo, and San Francisco, with diverse objects, environments, and motion patterns. Representative examples are shown in Fig.~\ref{fig:div}.



\clearpage
\section{Analysis under Occlusion and Depth Variation}
\label{supp:occ}
We analyze PECA under occlusion and depth variation using XYDeblur as a representative backbone. 
We estimate disparity with S\textsuperscript{2}M\textsuperscript{2}~\cite{min2025s2m2}
stereo matching on each rectified wide/ultra-wide pair and treat pixels with failed matching as occluded (hole) regions. These typically occur around object boundaries and depth discontinuities, where cross-view correspondence is unreliable. 
This fallback comes from the confidence-weighted residual fusion shared across backbones, so a single backbone suffices to analyze it.
Table~\ref{tab:occ} compares the single-view baseline, naive stereo concatenation, and PECA along three axes.

\vspace{-2mm}
\subsubsection{Region-wise.}
In non-occluded regions, PECA achieves a larger gain over the single-view baseline (+1.48 dB), indicating that it effectively exploits the reference view when correspondence is reliable. 
In occluded regions, the gain becomes smaller but remains positive (+0.20 dB), suggesting that PECA avoids severe degradation when reference cues are unreliable.

\vspace{-2mm}
\subsubsection{Hole ratio.}
Most images contain few hole pixels (mean 1.04\%, median 0.57\%, max 10.4\%), so we construct the most-occluded Top 30\% and Top 10\% subsets. 
Naive concatenation degrades as occlusion severity increases and falls below the single-view baseline on the Top 10\% subset ($-0.10$ dB). 
In contrast, PECA maintains positive gains across all subsets, including +0.32 dB on the Top 10\% subset.
Under unreliable conditions naive concatenation can be misled by incorrect reference information, whereas PECA achieves stable restoration through bounded search and residual fusion.

\vspace{-2mm}
\subsubsection{Relative depth.}
We group pixels by mean disparity into near, intermediate,
and far ranges, where PECA improves over the single baseline by +1.12 dB, +1.38 dB, and +1.59 dB. 
The lead over naive concatenation is largest in the near range, where larger disparity makes matching harder and the bounded directional search is most effective. 
At intermediate and far ranges the margin narrows, and most of the gain comes from the reference content itself.

\vspace{2mm}
\medskip\noindent
Overall, PECA exploits the reference view when reliable correspondence is available and remains stable when cross-view matching becomes unreliable. 
These results support that bounded epipolar attention and residual fusion suppress unreliable cross-view aggregation without requiring explicit occlusion masks.

\begin{table}[h]
\vspace{-3mm}
\scriptsize
\centering
\setlength{\tabcolsep}{5pt}
\caption{PSNR comparison on the XYDeblur backbone for occluded (OCC) and non-occluded (Non-OCC) regions, hole-severity subsets, and relative-depth subsets.}
\label{tab:occ}
\renewcommand{\arraystretch}{1.08}
\scalebox{0.9}{
\begin{tabular}{lcccccccc}
\toprule
& \multicolumn{2}{c}{Region-wise} & \multicolumn{3}{c}{Hole ratio} & \multicolumn{3}{c}{Relative depth} \\
\cmidrule(lr){2-3} \cmidrule(lr){4-6} \cmidrule(lr){7-9}
Method & OCC & Non-OCC & Full & Top 30\% & Top 10\% & Near & Intermediate & Far \\
\midrule
Single baseline 
& 29.74 
& 30.80 
& 30.76 
& 29.55 
& 28.03 
& 30.89
& 31.93  
& 31.34 \\

Stereo concatenation
& 29.78 
& 32.02 
& 31.96 
& 29.86 
& 27.93
& 31.68 	
& 33.28 
& \textbf{32.94} \\

\textbf{PECA (ours)}
& \textbf{29.94} 
& \textbf{32.28} 
& \textbf{32.22} 
& \textbf{30.53} 
& \textbf{28.35} 
& \textbf{32.01}	
& \textbf{33.31}   	
& 32.93\\
\bottomrule
\end{tabular}}
\vspace{-3mm}
\label{tab:occ}
\end{table}

\clearpage

\vspace{-3mm}
\section{Qualitative Comparison of PECA and Model Variants}
Figs.~\ref{fig:ch_st}-\ref{fig:ch_st1} present qualitative comparisons across three backbones, supporting the quantitative results in Table~\ref{tab:sota}.
\vspace{-2mm}

\subsubsection{Stereo information improves restoration quality.}
\begin{figure}[h]
  \centering
  \includegraphics[width=\linewidth]{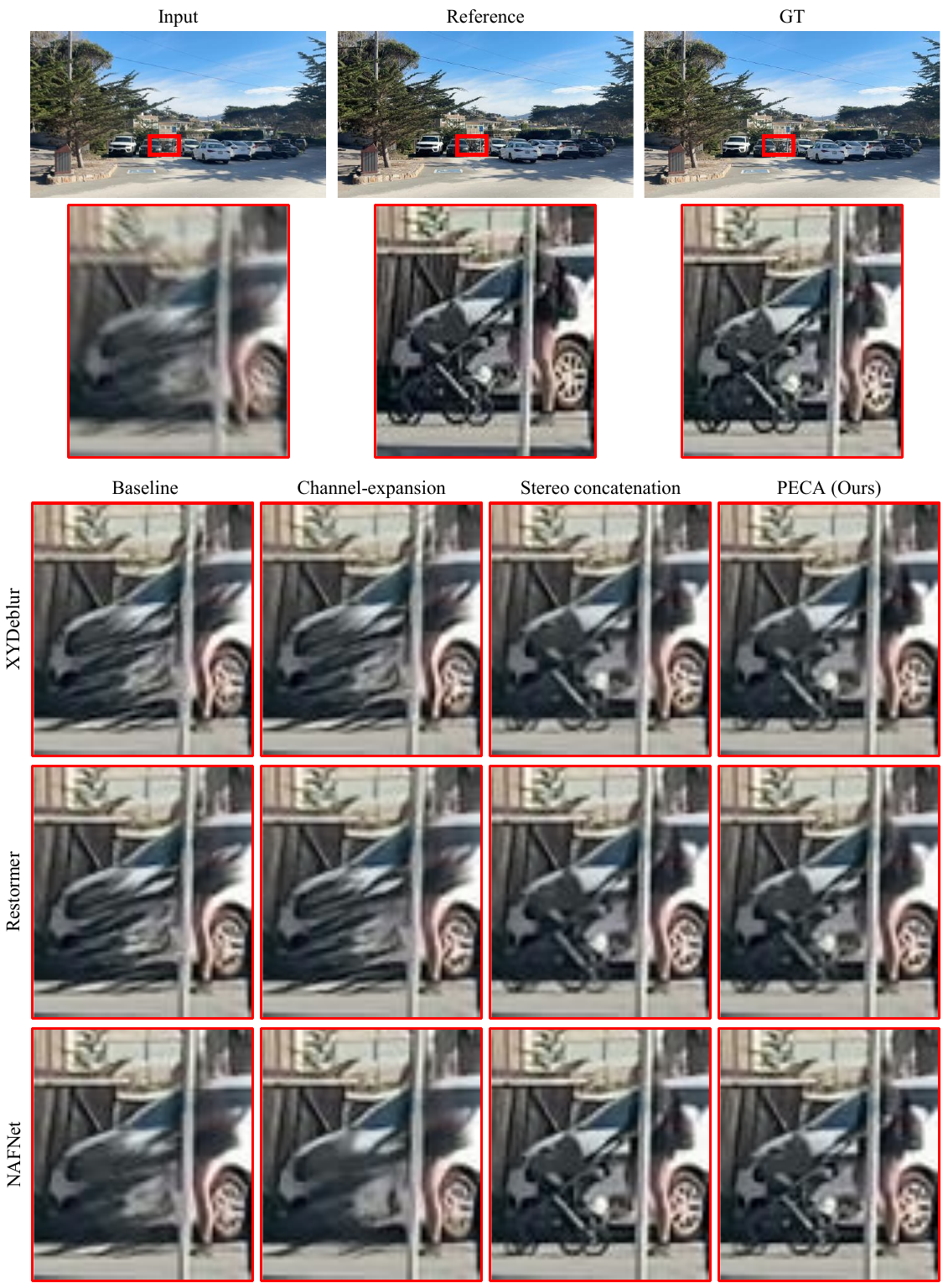}
  \caption{Qualitative results on the HSD dataset. A representative region is zoomed in to highlight large-scale motion blur and transparent structures around the vehicle.}
  \label{fig:ch_st}
\end{figure}

As shown in Fig.~\ref{fig:ch_st}, single-view settings (baseline and channel-expansion) are inherently limited under severe motion blur, where complex spatial layouts around the vehicle remain difficult to recover faithfully.
In contrast, stereo-view settings (stereo concatenation and PECA) leverage complementary cross-view information, enabling clearer reconstruction of object boundaries and overlapping structures.
\vspace{-2mm}

\subsubsection{Naive stereo fusion provides limited improvement.}
\begin{figure}[h]
  \centering
  \includegraphics[width=0.9\linewidth]{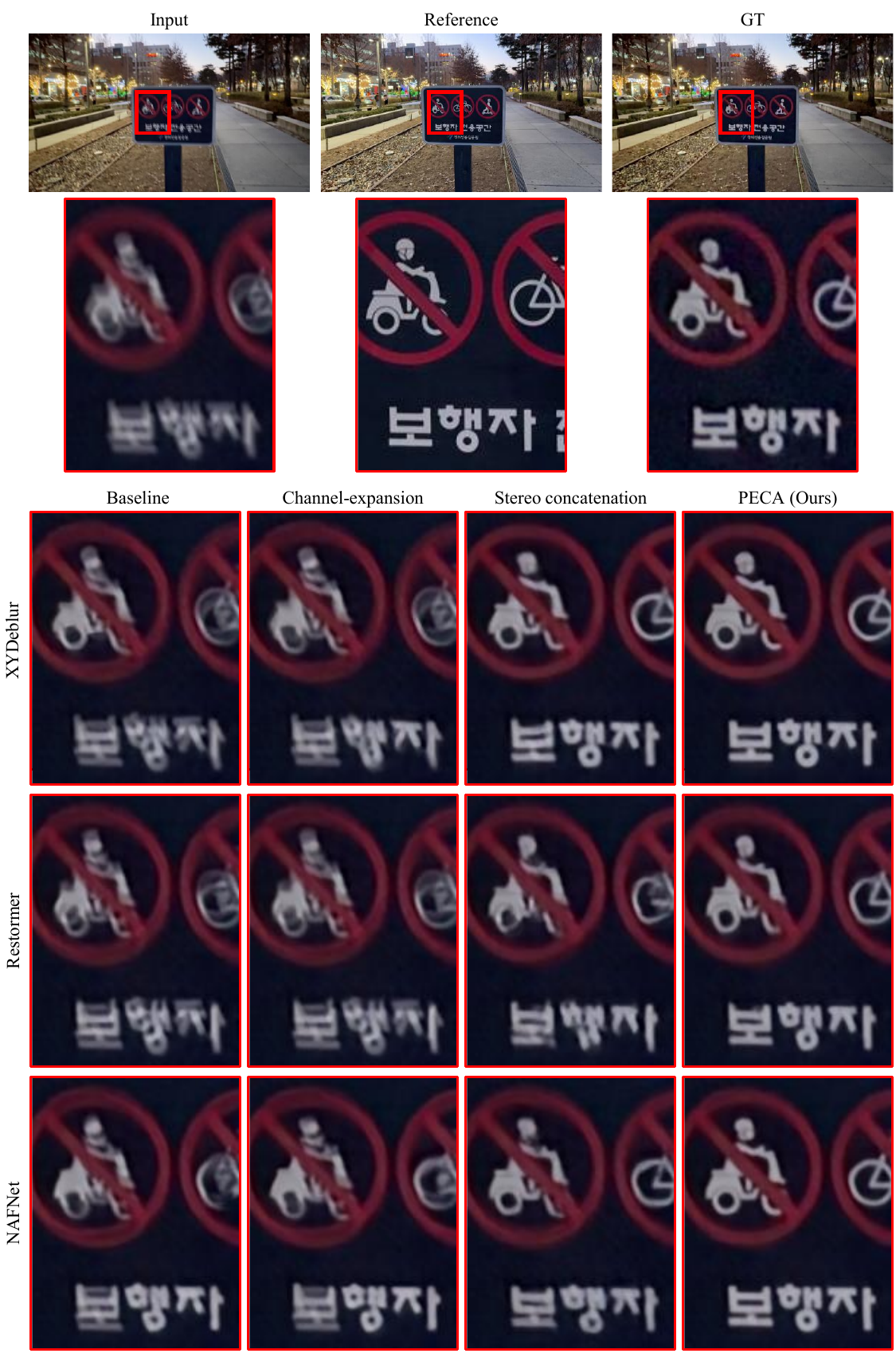}
  \caption{Qualitative results on the HSD dataset. Two regions are zoomed in for comparison, highlighting fine text and layered paper textures.} 
  \label{fig:ch_st2}
\end{figure}

While naive stereo concatenation can outperform single-view settings, its ability to exploit cross-view information remains limited.
In Fig.~\ref{fig:ch_st2}, stereo concatenation partially improves the restoration quality compared with single-view variants, yet still fails to recover fine structures.
In contrast, PECA produces noticeably clearer results by establishing more reliable cross-view correspondences.
\vspace{-2mm}

\subsubsection{Naive stereo fusion can even degrade performance.}
\begin{figure}[h]
  \centering
  \includegraphics[width=\linewidth]{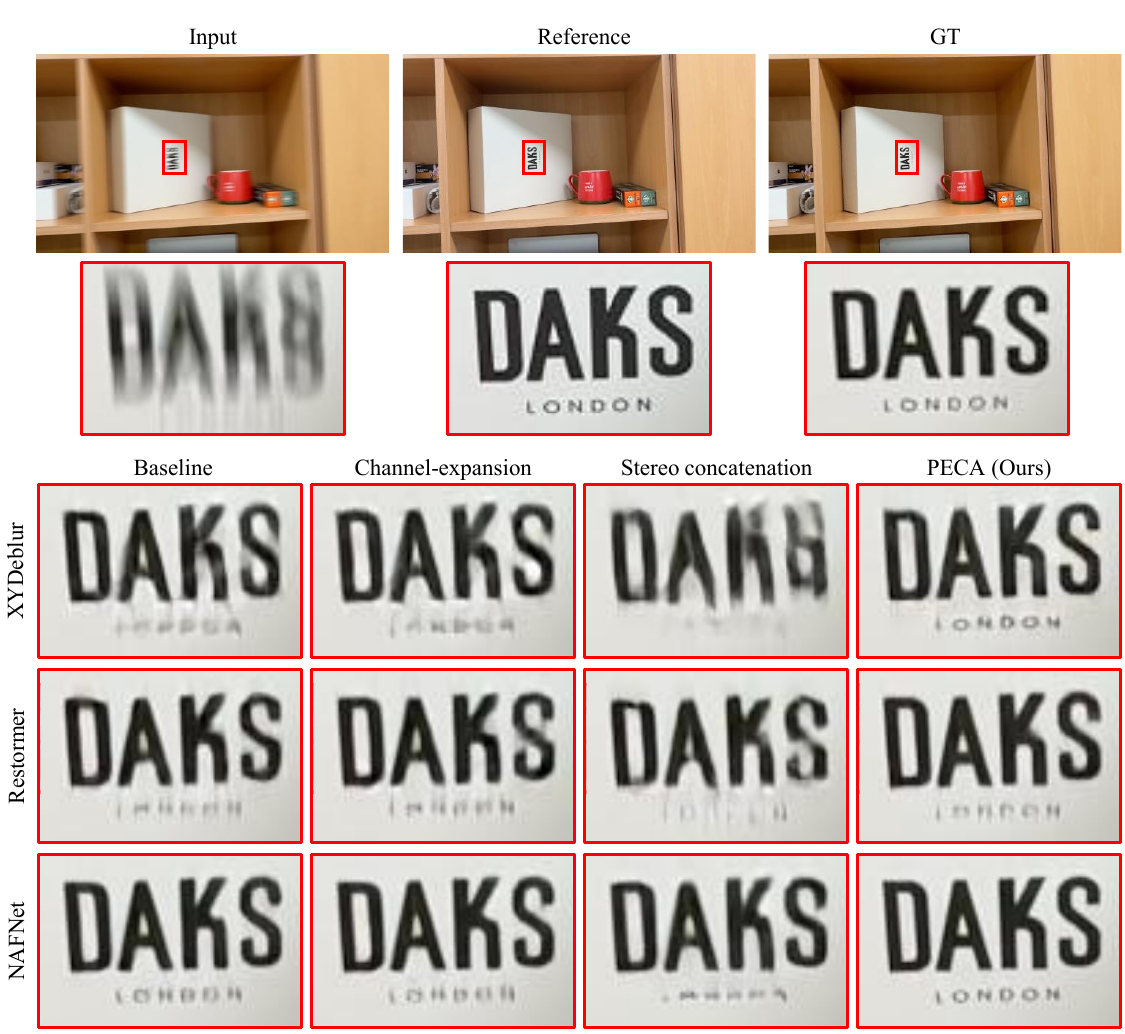}
  \caption{Qualitative results on the HSD dataset. A representative region is zoomed in to highlight fine-grained text on the box.}
  \label{fig:ch_st1}
\end{figure}

However, the effectiveness of stereo information depends strongly on the interaction mechanism.
As illustrated in Fig.~\ref{fig:ch_st1}, naive stereo concatenation does not consistently outperform single-view variants and may even yield inferior results compared to channel-expansion.
In contrast, PECA maintains robust restoration performance across all backbones.
By constraining the correspondence search space to a geometrically valid disparity range, PECA suppresses ambiguous matches and enables more accurate cross-view feature aggregation.

\clearpage

\section{Comparison with Recent Stereo Fusion Methods}

We conduct a controlled comparison by replacing only the stereo fusion module while keeping the backbone fixed.
As shown in Table~\ref{tab:sota_fusion}, PECA consistently outperforms stereo fusion modules such as SPAM~\cite{zhang2024stereo} and ALM~\cite{xiaoasymmetric} across all three backbones.
ALM incurs high computational cost due to dense alignment and falls below stereo concatenation on XYDeblur and Restormer, suggesting that dense matching can be less suitable for asymmetric deblurring. SPAM restricts matching to epipolar scanlines but lacks an explicit physical disparity bound. In contrast, PECA enforces both the disparity direction and a bounded search range, reducing ambiguous candidates under asymmetric blur. As a result, PECA achieves the best restoration accuracy with favorable module complexity.

\vspace{-5mm}
\begin{table}[h]
\scriptsize
\centering
\setlength{\tabcolsep}{5pt}
\caption{Quantitative comparison with stereo fusion modules}
\vspace{-2mm}
\label{tab:sota_fusion}
\renewcommand{\arraystretch}{1.08}
\scalebox{0.9}{
\begin{tabular}{lcccccccc}
\toprule
& \multicolumn{2}{c}{XYDeblur~\cite{ji2022xydeblur}} 
& \multicolumn{2}{c}{Restormer~\cite{zamir2022restormer}} 
& \multicolumn{2}{c}{NAFNet~\cite{chen2022simple}} 
& \multicolumn{2}{c}{Module} \\
\cmidrule(lr){2-3} \cmidrule(lr){4-5}
\cmidrule(lr){6-7} \cmidrule(lr){8-9}
Method 
& PSNR & SSIM 
& PSNR & SSIM 
& PSNR & SSIM 
& MACs(G) & Params(K) \\
\midrule

Stereo concatenation
& 31.95 & 0.9590 
& 31.83 & 0.9581 
& 32.23 & 0.9601 
& - & - \\

SPAM~\cite{zhang2024stereo}
& 32.02 & 0.9601 
& 32.04 & 0.9580 
& 32.24 & 0.9612 
& 11.3 & 188.6 \\

ALM~\cite{xiaoasymmetric}
& 30.81 & 0.9448 
& 31.07 & 0.9460 
& 32.05 & 0.9574 
& 426.6 & 32.9 \\


\textbf{PECA (ours)}
& \textbf{32.22} & \textbf{0.9622} 
& \textbf{32.47} & \textbf{0.9636} 
& \textbf{32.92} & \textbf{0.9669} 
& \textbf{4.8} & \textbf{83.1} \\

\bottomrule
\end{tabular}}
\vspace{-3mm}
\end{table}

\vspace{-5mm}
\section{Qualitative Cross-device Evaluation}
\label{supp:galaxy}

\begin{figure}[h]
  \centering
  \includegraphics[width=0.85\linewidth]{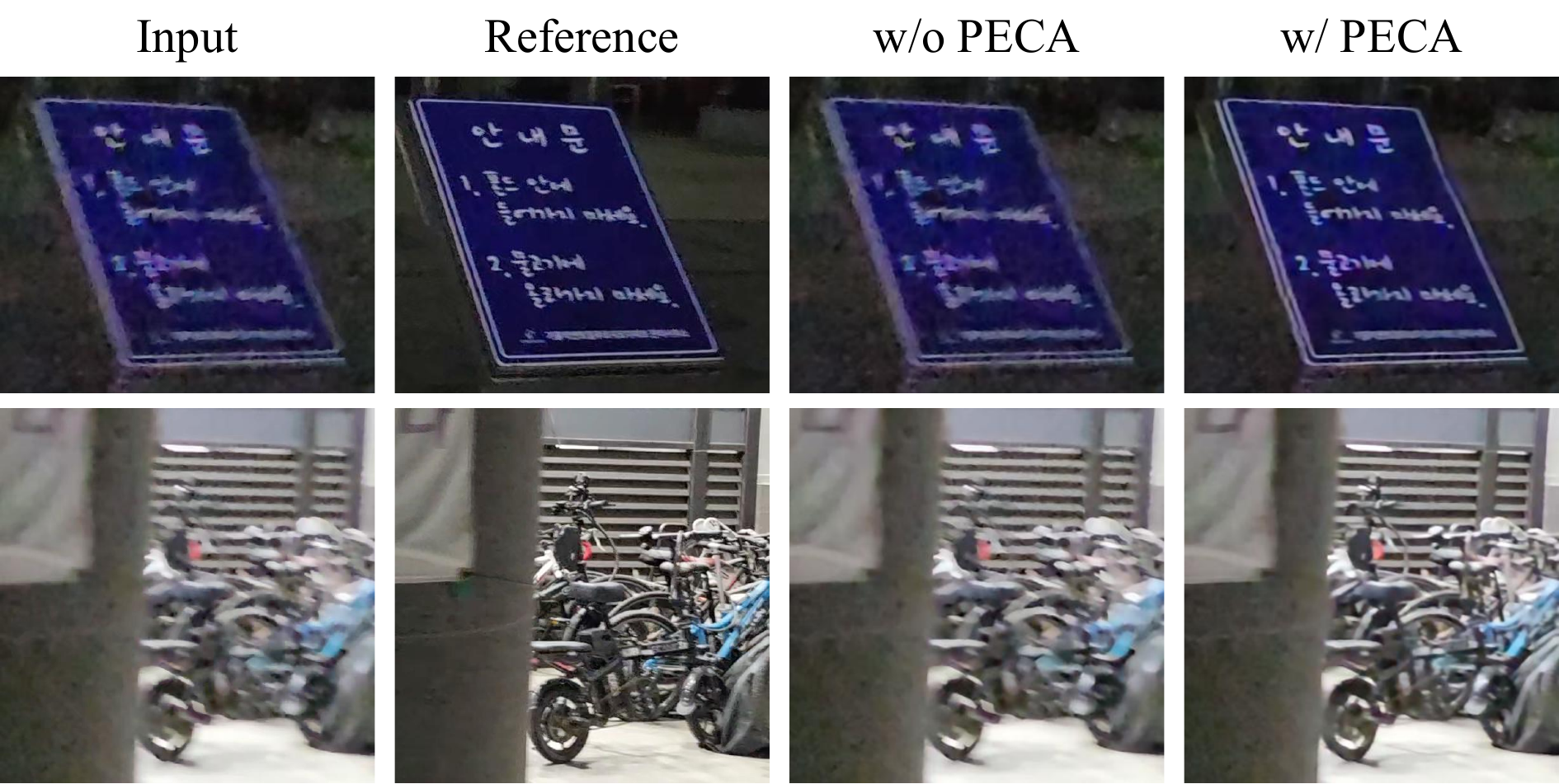}
    \vspace{-2mm}
  \caption{Qualitative results on real blur captured by Galaxy S25.}
  \label{fig:s25}
\end{figure}

To further assess the cross-device generalization of PECA, we provide additional qualitative results on real-blur examples captured by an unseen Galaxy smartphone. Fig.~\ref{fig:s25} shows that PECA yields clearer restoration than the baseline without PECA. Specifically, PECA better preserves the boundary and improves the legibility of text details in the first example, and recovers sharper object structures around the bicycles in the second example.
While a comprehensive quantitative cross-device evaluation remains future work, these results suggest that PECA generalizes to real blur from unseen devices.

\vspace{-3mm}

\clearpage

\begin{figure}[t]
\centering
\includegraphics[width=\linewidth]{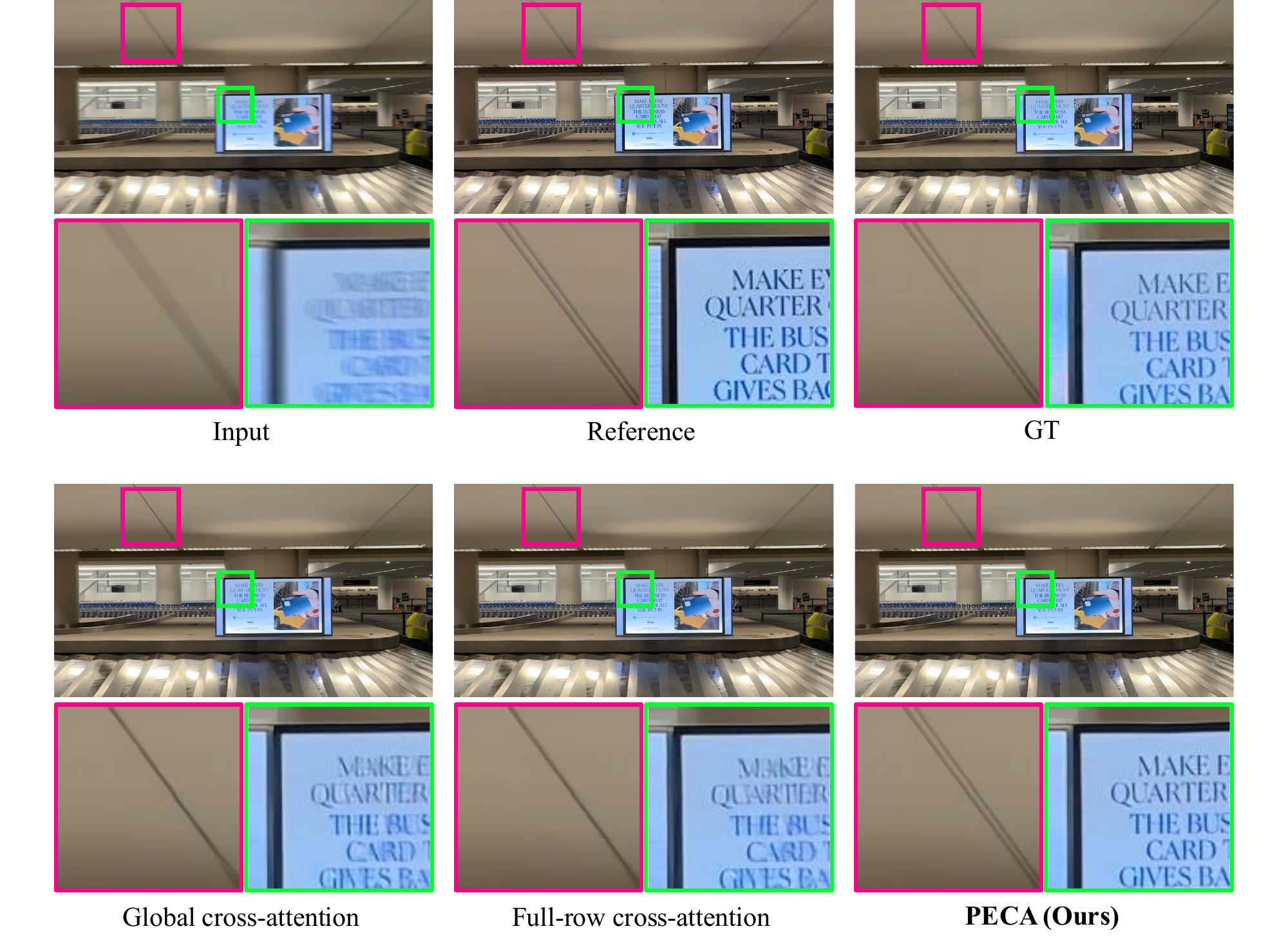}
\caption{Qualitative comparison of global cross attention, full-row cross attention, and PECA. 
Two representative regions are magnified for comparison, highlighting the ceiling edge and the fine text on the signboard.}
\label{fig:window}
\end{figure}

\section{Effect of Attention Search Ranges}
\label{supp:search_range}
Fig.~\ref{fig:window} presents qualitative results corresponding to the quantitative comparison in Table~\ref{tab:search_range}.
The highlighted regions illustrate challenging cases where conventional cross attention mechanisms fail to restore fine details under severe blur.
In particular, the results show that an excessively wide correspondence search range can degrade feature aggregation in different ways.
For example, it may dilute a plausible dominant match by mixing weak responses from other locations, or it may induce multi-modal matching ambiguity when repetitive fine structures produce several competing candidates.

In the red-highlighted region of Fig.~\ref{fig:window}, the input contains two thin parallel line structures. 
Both global attention and full-row attention collapse these two lines into a single thicker edge.
Under severe blur, the responses of the two lines become broadened and less sharply localized.
When the search range is too wide, the number of weakly relevant or non-corresponding candidates increases substantially. Although each such location may receive only a small attention weight, their accumulated contribution becomes non-negligible during value aggregation and dilutes the dominant correspondence peak.
As a result, the matched value feature no longer dominates the aggregation, and the reconstructed feature becomes less selective and more averaged. 
This over-smoothed aggregation suppresses the high-frequency separation between the two lines and biases the reconstruction toward lower-frequency content, causing the two thin lines to merge into a thicker structure.
In contrast, PECA constrains the correspondence search to a geometrically valid one-sided disparity interval that not only restricts the overall search range but also reduces the candidate region over which blur-broadened responses can spread. This decreases the number of irrelevant candidates that contribute to peak dilution and better preserves the separation between the two lines.

In the green-highlighted region, a different failure mode appears in fine textual patterns. 
Here, severe blur attenuates high-frequency stroke details, while repeated and visually similar character strokes generate multiple plausible correspondence candidates. 
With a wide search range, attention becomes multi-modal and aggregates features from several competing locations rather than selecting a single reliable match. 
This ambiguity mixes distinct stroke patterns, leading to distorted characters and reduced readability. 
By constraining the correspondence search space, PECA suppresses these ambiguous responses and enables more stable correspondence selection, resulting in clearer stroke boundaries.

These qualitative observations are consistent with the quantitative improvements reported in Table~\ref{tab:search_range}. 
These results demonstrate that constraining the correspondence search space is critical for reliable cross-view deblurring under asymmetric blur conditions.




\section{Analysis of Cross-view Correspondence in PECA}
\label{supp:rel_pos}

We analyze the difference between PECA and prior 1D epipolar attention mechanisms. 
Existing full-row epipolar attention methods compute correspondence over the entire scanline, which may include implausible candidates.
In contrast, PECA jointly enforces the rectified-stereo direction and a bounded disparity range.

\vspace{-4mm}
\begin{figure}[h]
  \centering
  \includegraphics[width=0.9\linewidth]{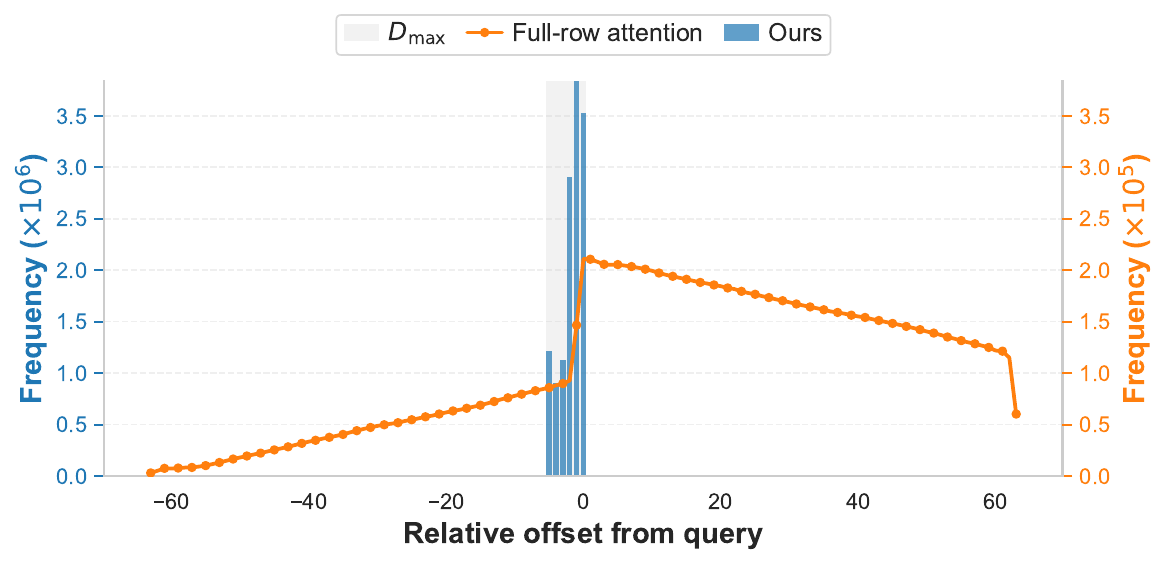}
  \vspace{-3mm}
  \caption{Top-3 attention match distribution along the epipolar line.}
  \label{fig:supp_rel_pos_top3}
\end{figure}

\vspace{-4mm}
Fig.~\ref{fig:supp_rel_pos_top3} visualizes the distribution of the top-3 attention positions relative to each query. Full-row attention spreads attention across the entire scanline, including large positive and negative offsets, whereas PECA concentrates its attention within the bounded disparity range. This result explains the performance gap between full-row attention and PECA.

\section{Qualitative Comparison on $D_{\max}$ and $\tau$}
\label{supp:hyper}
\begin{figure}[h]
  \centering
  \includegraphics[width=\linewidth]{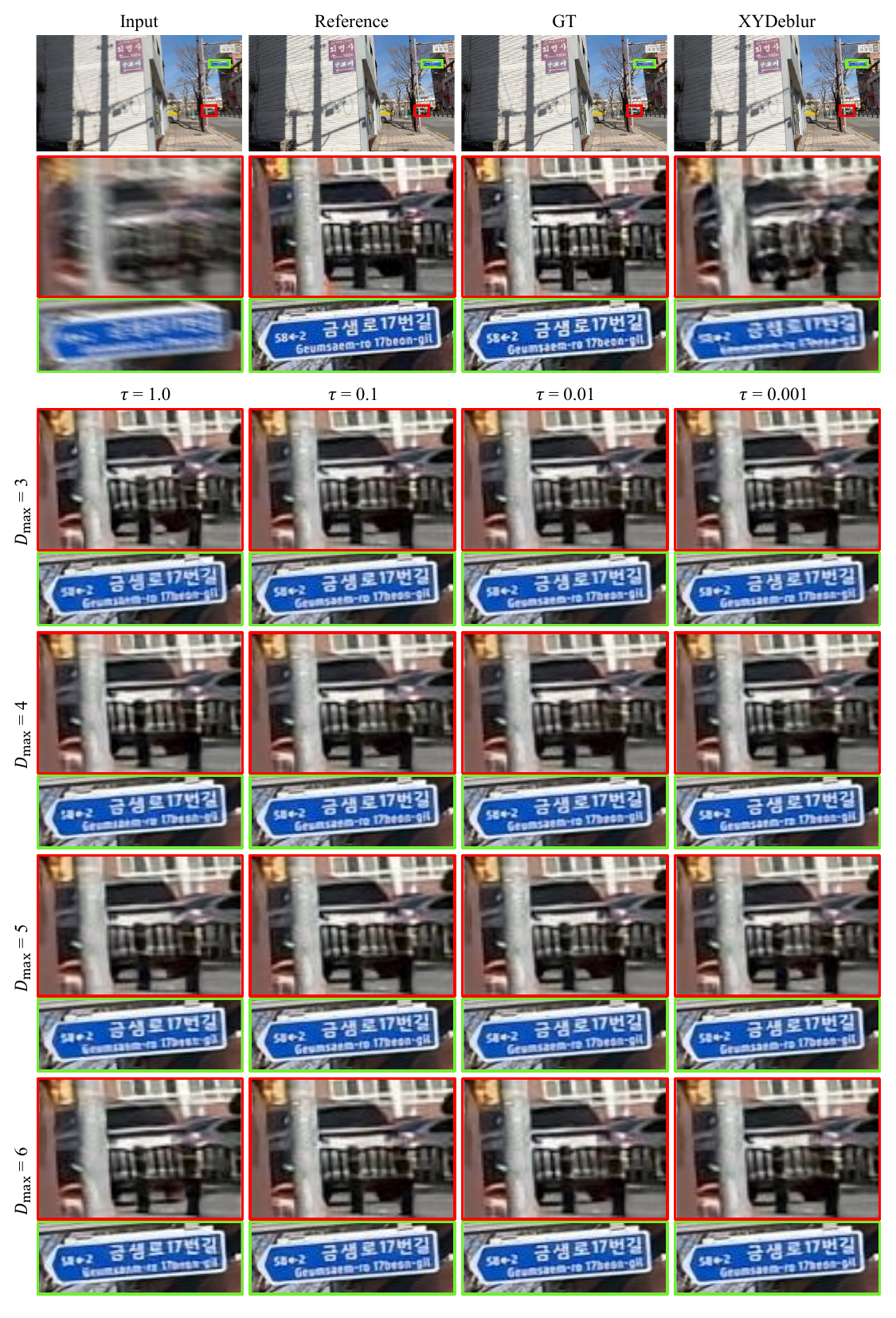}
  \vspace{-5mm}
  \caption{Qualitative results on the HSD dataset for $D_{\max}=3$–$6$ under different softmax temperatures $\tau$.}
  \label{fig:hyper_xy}
\end{figure}

\begin{figure}[h]
  \centering
  \includegraphics[width=\linewidth]{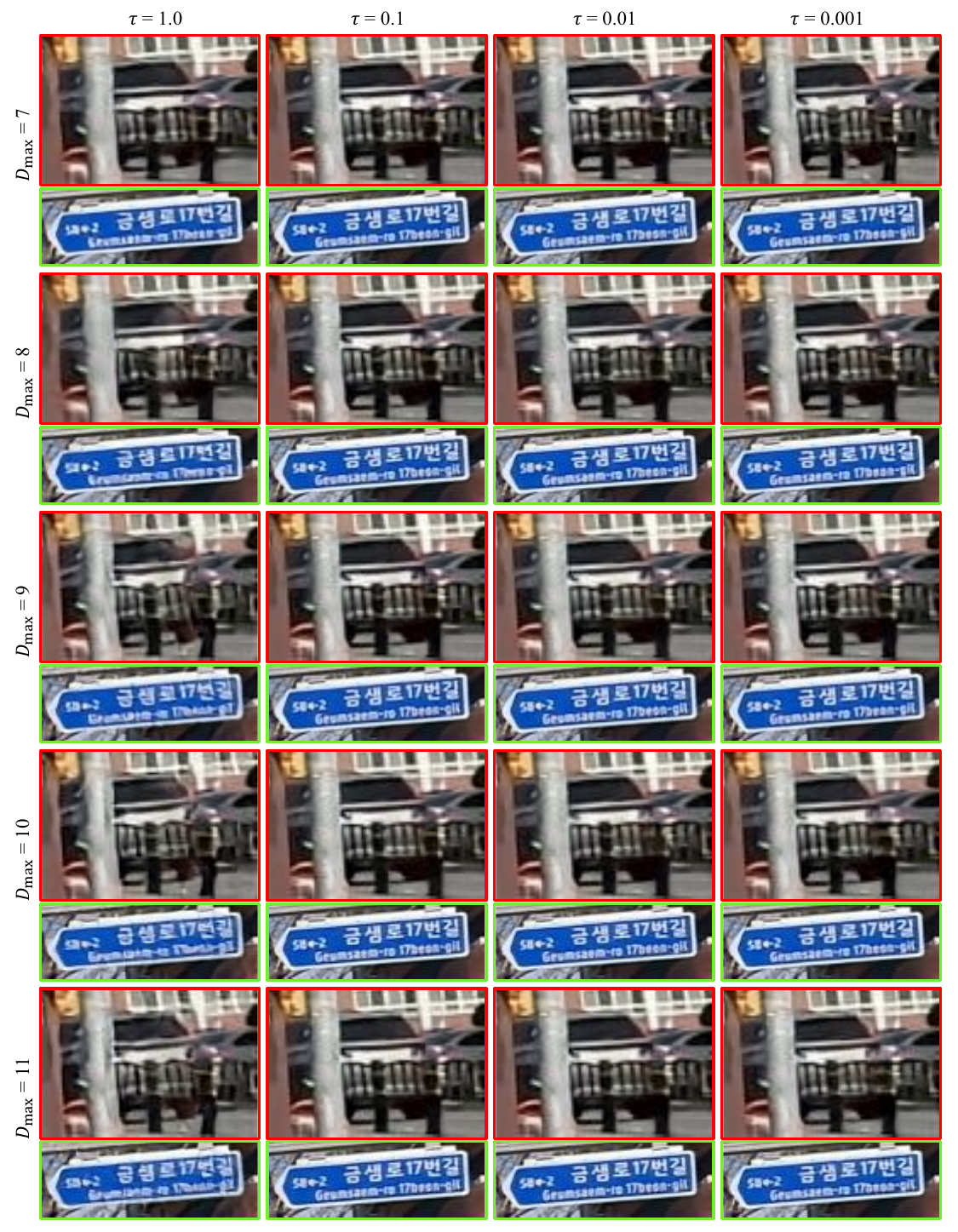}
  \vspace{-5mm}
  \caption{Qualitative results on the HSD dataset for $D_{\max}=7$–$11$ under different softmax temperatures $\tau$.}
  \label{fig:hyper_xy2}
\end{figure}

\begin{figure}[h]
  \centering
  \includegraphics[width=\linewidth]{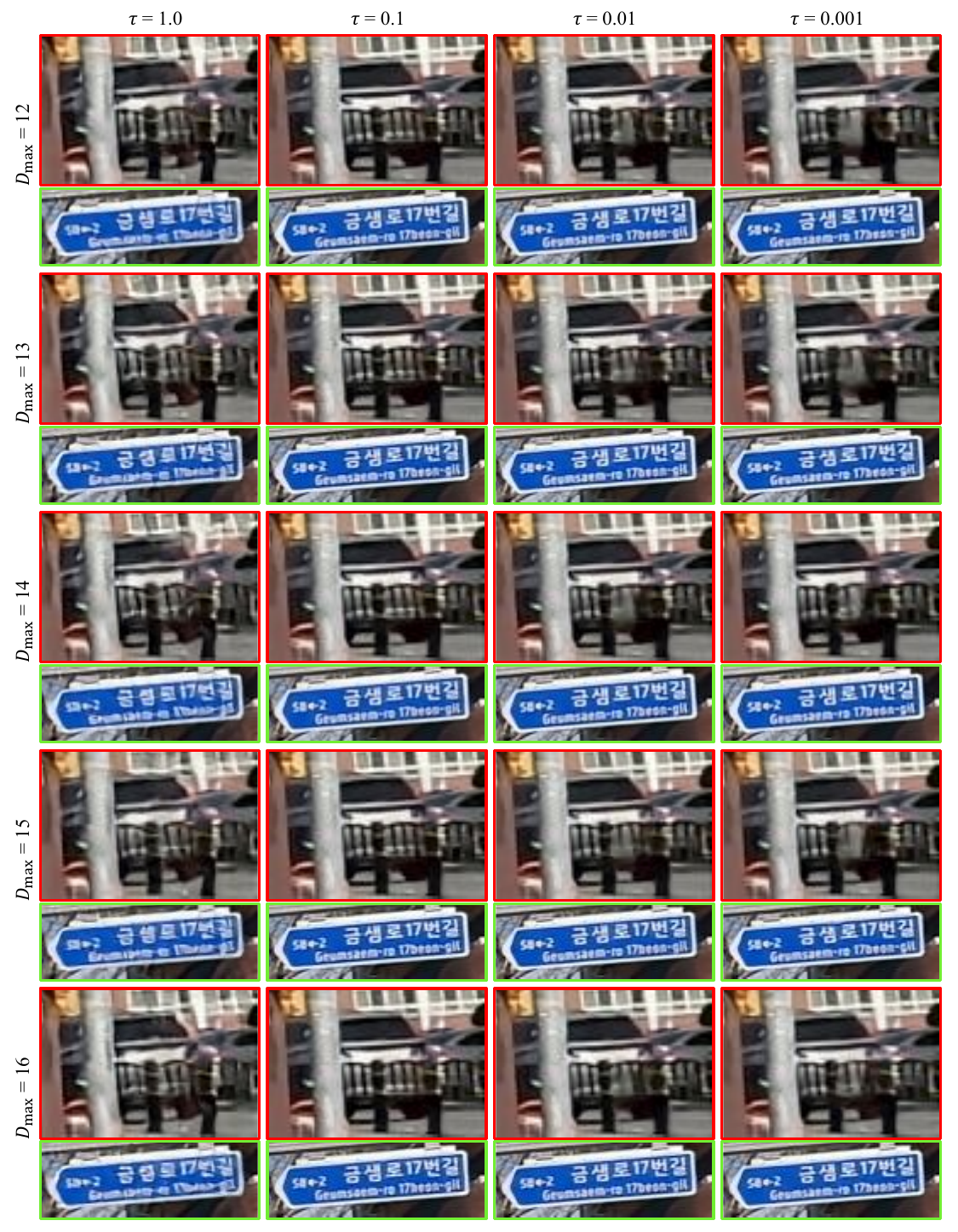}
  \vspace{-5mm}
  \caption{Qualitative results on the HSD dataset for $D_{\max}=12$–$16$ under different softmax temperatures $\tau$.}
  \label{fig:hyper_xy3}
\end{figure}

Figs.~\ref{fig:hyper_xy}-\ref{fig:hyper_xy3} present qualitative comparisons for different epipolar window sizes $D_{\max}$ and softmax temperatures $\tau \in \{1.0, 0.1, 0.01, 0.001\}$ on the HSD dataset using the XYDeblur backbone. Two zoomed regions highlight challenging cases, including large-scale motion blur and fine-grained text.

Increasing $D_{\max}$ enlarges the search space along the epipolar direction. 
Although a larger window allows PECA to consider a broader set of candidates, an excessively large $D_{\max}$ introduces redundant or ambiguous correspondences that can degrade the matching reliability. 
In such cases, irrelevant features from distant locations are less effectively suppressed, distracting attention from the true correspondence and leading to degraded restoration quality. 
Notably, when $D_{\max}$ becomes large (\eg, $D_{\max} \geq 12$), the foreground railings in front of the vehicle gradually disappear or merge with the background, as the enlarged search space introduces incorrect feature aggregation.

When the temperature is large (\eg, $\tau{=}1.0$), the attention distribution becomes overly smooth, causing the attention weights to spread across multiple candidates rather than concentrate on reliable correspondences. 
As a result, biased matches may arise from several locations along the scanline, which weakens the effectiveness of PECA and leads to blurry restorations.
In practice, $\tau{=}1.0$ yields inferior visual quality across all tested values of $D_{\max}$, producing blurrier results than smaller temperature values. This effect is particularly noticeable in structured regions such as the vehicle body and the text on the signboard.

\clearpage

\newcommand{\pos}[1]{\textcolor{blue!70!black}{\ensuremath{(+#1)}}}
\newcommand{\negat}[1]{\textcolor{red!75!black}{\ensuremath{(-#1)}}}

\begin{table}[h]
\centering
\setlength{\tabcolsep}{6pt}
\caption{PSNR comparison on different difficulty subsets of the test set.}
\label{tab:subset}
\resizebox{0.9\linewidth}{!}{
\begin{tabular}{lcccccc}
\toprule
\textbf{Method} & \multicolumn{6}{c}{\textbf{PSNR $ \uparrow$}} \\
\cmidrule(lr){2-7}
 & \multicolumn{2}{c}{Easy ($\Delta$)} & \multicolumn{2}{c}{Moderate ($\Delta$)} & \multicolumn{2}{c}{Hard ($\Delta$)} \\
\midrule
XYDeblur~\cite{ji2022xydeblur}   & 34.38 & - & 31.51 & - & 25.65 & - \\
\quad + Global cross attention & 34.37 & \negat{0.01} & 31.45 & \negat{0.06} & 25.62 & \negat{0.03} \\
\quad + Full-row cross attention & 34.40 & \pos{0.02} & 31.49 & \negat{0.02} & 25.62 & \negat{0.03} \\
\quad + Stereo concatenation  & 34.80 & \pos{0.42} & 32.73 & \pos{1.22} & 27.58 & \pos{1.93} \\
\quad + \textbf{PECA (Ours)}  & \textbf{34.88} & \pos{0.50} & \textbf{32.98} & \pos{1.47} & \textbf{28.03} & \pos{2.38} \\
\midrule
Restormer~\cite{zamir2022restormer}  & 34.73 & - & 31.68 & - & 25.66 & - \\
\quad + Stereo concatenation & 34.98 & \pos{0.25} & 32.48 & \pos{0.80} & 27.37 & \pos{1.71} \\
\quad + \textbf{PECA (Ours)}  & \textbf{35.45} & \pos{0.72} & \textbf{33.08} & \pos{1.40} & \textbf{28.26} & \pos{2.60} \\
\midrule
NAFNet~\cite{chen2022simple}     & 35.22 & - & 32.88 & - & 27.66 & - \\
\quad + Stereo concatenation & 35.23 & \pos{0.01} & 33.08 & \pos{0.20} & 27.23 & \negat{0.43} \\
\quad + \textbf{PECA (Ours)}  & \textbf{35.36} & \pos{0.14} & \textbf{33.66} & \pos{0.78} & \textbf{28.99} & \pos{1.33} \\
\bottomrule
\end{tabular}
}
\end{table}

\begin{table}[h]
\centering
\setlength{\tabcolsep}{6pt}
\caption{SSIM comparison on different difficulty subsets of the test set.}
\label{tab:subset_ssim}
\resizebox{0.9\linewidth}{!}{
\begin{tabular}{lcccccc}
\toprule
\textbf{Method} & \multicolumn{6}{c}{\textbf{SSIM $\uparrow$}} \\
\cmidrule(lr){2-7}
 & \multicolumn{2}{c}{Easy ($\Delta$)} & \multicolumn{2}{c}{Moderate ($\Delta$)} & \multicolumn{2}{c}{Hard ($\Delta$)} \\
\midrule
XYDeblur~\cite{ji2022xydeblur}   & 0.9811 & - & 0.9626 & - & 0.8714 & - \\
\quad + Global cross attention & 0.9810 & \negat{0.0001} & 0.9622 & \negat{0.0004} & 0.8702 & \negat{0.0012} \\
\quad + Full-row cross attention & 0.9809 & \negat{0.0002} & 0.9620 & \negat{0.0006} & 0.8702 & \negat{0.0012} \\
\quad + Stereo concatenation  & 0.9829 & \pos{0.0018} & 0.9717 & \pos{0.0091} & 0.9118 & \pos{0.0404} \\
\quad + \textbf{PECA (Ours)}  & \textbf{0.9830} & \pos{0.0019} & \textbf{0.9729} & \pos{0.0103} & \textbf{0.9198} & \pos{0.0484} \\
\midrule
Restormer~\cite{zamir2022restormer}  & 0.9817 & - & 0.9625 & - & 0.8709 & - \\
\quad + Stereo concatenation & 0.9826 & \pos{0.0009} & 0.9688 & \pos{0.0063} & 0.9119 & \pos{0.0410} \\
\quad + \textbf{PECA (Ours)}  & \textbf{0.9838} & \pos{0.0021} & \textbf{0.9723} & \pos{0.0098} & \textbf{0.9260} & \pos{0.0551} \\
\midrule
NAFNet~\cite{chen2022simple}     & 0.9834 & - & 0.9702 & - & 0.9077 & - \\
\quad + Stereo concatenation & 0.9837 & \pos{0.0003} & 0.9725 & \pos{0.0023} & 0.9125 & \pos{0.0048} \\
\quad + \textbf{PECA (Ours)}  & \textbf{0.9841} & \pos{0.0007} & \textbf{0.9753} & \pos{0.0051} & \textbf{0.9327} & \pos{0.0250} \\
\bottomrule
\end{tabular}
}
\end{table}

\vspace{-5mm}
\section{Performance Analysis on Test Subsets}

To evaluate performance across different levels of degradation, we partition the test set into three difficulty-based subsets according to the PSNR between blurry and sharp image pairs.
Specifically, samples in the bottom $25\%$ (PSNR \mbox{$< 24.4$ dB}) are categorized as Hard, those in the middle $50\%$ (24.4--30.3 dB) as Moderate, and those in the top $25\%$ (PSNR $> 30.3$ dB) as Easy.
These subsets contain 275, 550, and 275 image pairs, respectively.


Table~\ref{tab:subset}, ~\ref{tab:subset_ssim} report quantitative results on the difficulty-based subsets. PECA consistently outperforms all compared variants across all subsets in terms of both PSNR and SSIM. Notably, the performance gains become more pronounced on the Hard subset, where severe blur and complex scene structures make correspondence estimation more challenging.
Compared with stereo concatenation, PECA improves PSNR on the Hard subset by 0.45 dB for XYDeblur, 0.89 dB for Restormer, and 1.76 dB for NAFNet.
Similar trends can be observed in SSIM, with clear improvements on the most challenging samples across all three backbones. 
These results indicate that PECA can effectively exploit cross-view information under severe degradation, where naive stereo fusion or unconstrained matching becomes less reliable. Moreover, the results on XYDeblur show that global or full-row cross attention may even underperform the single-view baseline, further highlighting the importance of constraining the correspondence search space.

\end{document}